\documentclass[pdflatex,sn-mathphys-num]{sn-jnl}

\usepackage{graphicx}%
\usepackage{multirow}%
\usepackage{amsmath,amssymb,amsfonts}%
\usepackage{amsthm}%
\usepackage{mathrsfs}%
\usepackage[title]{appendix}%
\usepackage{xcolor}%
\usepackage{textcomp}%
\usepackage{manyfoot}%
\usepackage{booktabs}%
\usepackage{algorithm}%
\usepackage{algorithmicx}%
\usepackage{algpseudocode}%
\usepackage{listings}%
\usepackage{rotating}
\usepackage{array}

\theoremstyle{thmstyleone}%
\theoremstyle{thmstyletwo}%

\theoremstyle{thmstylethree}%

\begin{document}

\title[Article Title]{DPWMixer: Dual-Path Wavelet Mixer for Long-Term Time Series Forecasting}


\author[1]{\fnm{Qianyang} \sur{Li}}\email{liqianyang@stu.xjtu.edu.cn}
\author*[1]{\fnm{Xingjun} \sur{Zhang}}\email{xjzhang@xjtu.edu.cn}
\author[1]{\fnm{Shaoxun} \sur{Wang}}\email{shaoxunwang@stu.xjtu.edu.cn}
\author[2]{\fnm{Jia} \sur{Wei}}\email{weijia4473@mail.tsinghua.edu.cn}

\affil*[1]{\orgdiv{School of Computer Science and Technology}, \orgname{Xi'an Jiaotong University}, \orgaddress{ \city{Xi’an} \postcode{710049}, \country{China}}}
\affil[2]{\orgdiv{Department of Computer Science and Technology}, \orgname{Tsinghua University},  \orgaddress{ \city{Beijing} \postcode{100084}, \country{China}}}


\abstract{
Long-term time series forecasting (LTSF) is a critical task in computational intelligence. While Transformer-based models effectively capture long-range dependencies, they often suffer from quadratic complexity and overfitting due to data sparsity. Conversely, efficient linear models struggle to depict complex non-linear local dynamics. Furthermore, existing multi-scale frameworks typically rely on average pooling, which acts as a non-ideal low-pass filter, leading to spectral aliasing and the irreversible loss of high-frequency transients. In response, this paper proposes DPWMixer, a computationally efficient Dual-Path architecture. The framework is built upon a Lossless Haar Wavelet Pyramid that replaces traditional pooling, utilizing orthogonal decomposition to explicitly disentangle trends and local fluctuations without information loss. To process these components, we design a Dual-Path Trend Mixer that integrates a global linear mapping for macro-trend anchoring and a flexible patch-based MLP-Mixer for micro-dynamic evolution. Finally, An adaptive multi-scale fusion module then integrates predictions from diverse scales, weighted by channel stationarity to optimize synthesis. Extensive experiments on eight public benchmarks demonstrate that our method achieves a consistent improvement over state-of-the-art baselines. The code is available at https://github.com/hit636/DPWMixer.}




\maketitle

\section{Introduction}\label{sec1}

Time series forecasting (TSF) is fundamental for data-driven decision-making systems, including energy grid management \cite{deb2017review,LAGO2021116983}, intelligent transportation systems \cite{abirami2024systematic,cirstea2022towards}, and climate modeling \cite{wong2024ai}.Recently, researchers have started to investigate Long-Term Time Series Forecasting (LTSF) \cite{qiu2024tfb,kim2025comprehensive}, where models are asked to forecast substantial future horizons given historical context. Given the computationally demanding nature of the task, models are challenged by the involved temporal dynamics, which are characterized by non-stationarity, multi-scale periodicity, and the superposition of deterministic trends with stochastic noise.

Approaches have evolved from recurrent structures to Transformer-based models \cite{wen2022transformers,vaswani2017attention}. To address self-attention issues for long sequences, methods such as Informer \cite{zhou2021informer}, Autoformer \cite{wu2021autoformer}, and PatchTST \cite{nie2022time} have been developed to alleviate quadratic computational complexity and to improve local semantic extraction. These methods are able to model long-range dependencies, they tend to be memory-intensive and are prone to overfitting. This is an issue due to the sparse semantic density of time series data, compared to natural language.
On the other hand, Multi-Layer Perceptron (MLP) and linear-based models, such as DLinear \cite{zeng2023transformers} and TiDE \cite{DBLP:journals/tmlr/DasKLMSY23}, have shown that simple models can obtain competitive results with linear complexity $\mathcal{O}(L)$. However, the use of static weights in these models hinders their ability to model the non-linear evolutionary dynamics in rapidly changing distributions.

Despite these advancements, modeling real-world LTSF data presents two fundamental challenges derived from the intrinsic characteristics of time series signals, as illustrated in Figure \ref{fig:motivation}:
\begin{figure*}[h]
    \centering 
    \begin{minipage}[b]{0.48\columnwidth}
        \centering
        \includegraphics[width=\linewidth]{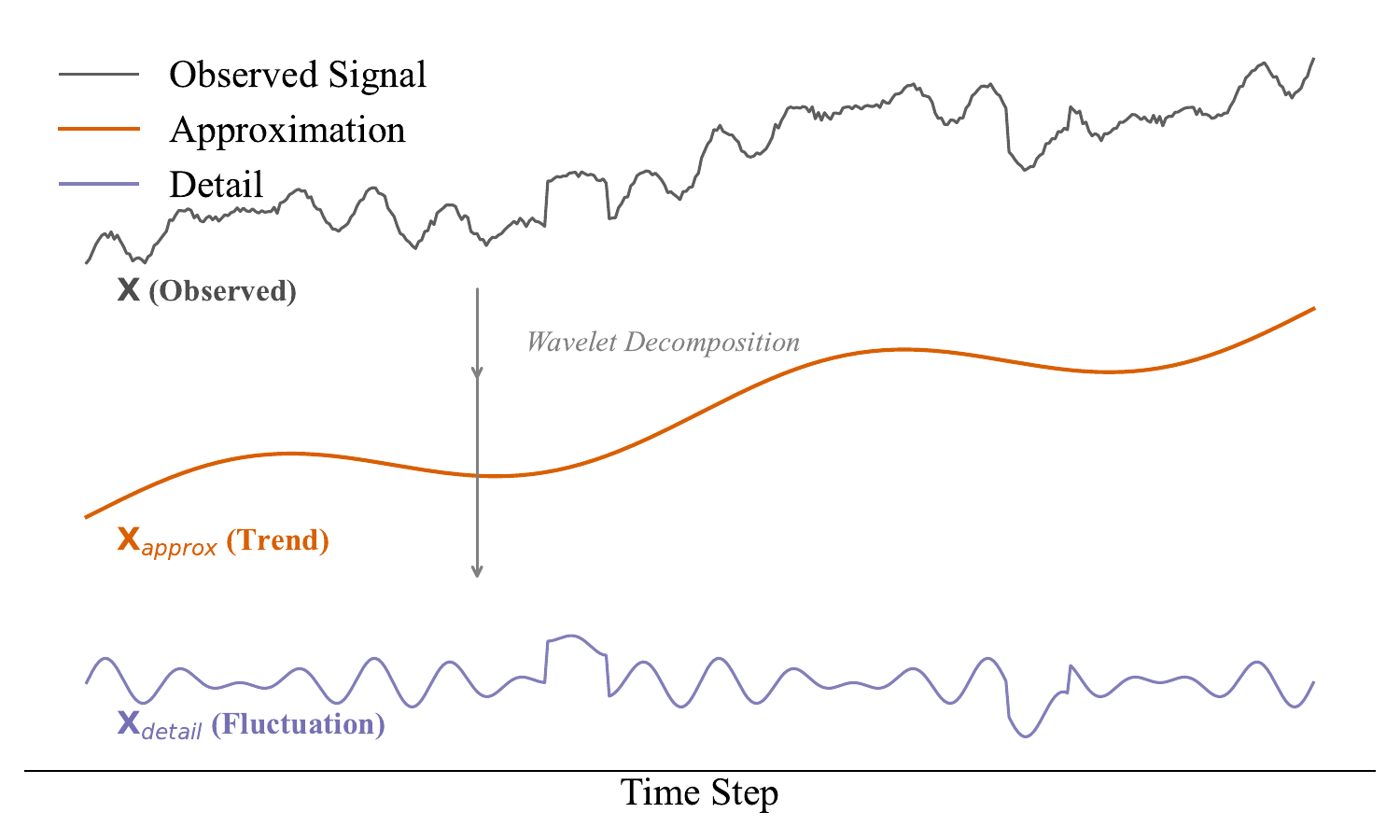}
        \centerline{(a) Spectral Aliasing (Decomposition)}
    \end{minipage}%
    \hfill
    \begin{minipage}[b]{0.48\columnwidth}
        \centering
        \includegraphics[width=\linewidth]{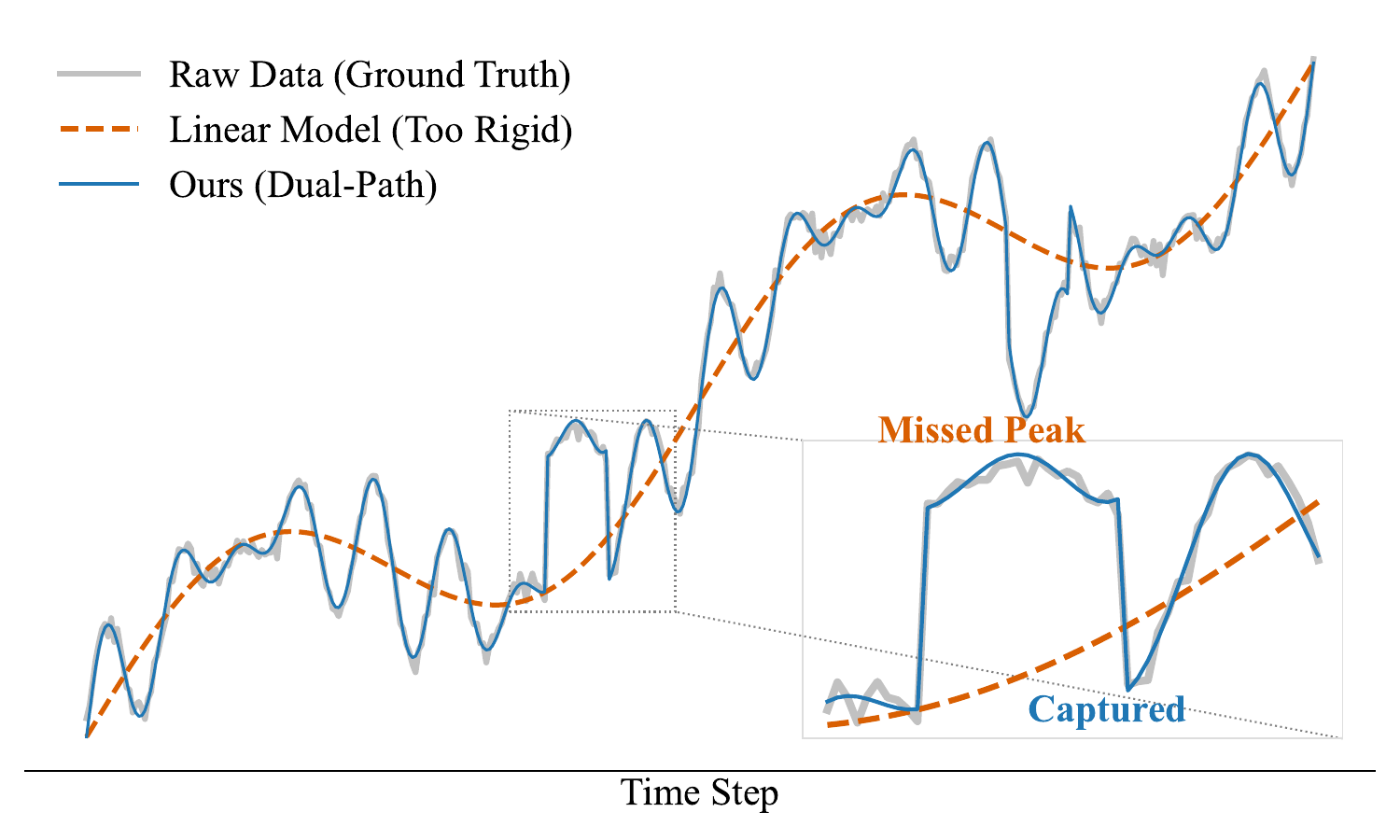}
        \centerline{(b) Trend-Detail Incompatibility (Modeling)}
    \end{minipage}   
   \caption{Illustration of the intrinsic challenges in LTSF and our design intuition. 
   \textbf{(a) Spectral Aliasing (Decomposition):} Standard pooling leads to aliasing and information loss, whereas our wavelet-based approach achieves lossless, orthogonal disentanglement.
    \textbf{(b) Trend-Detail Incompatibility (Modeling):} Pure linear models (Orange dashed) capture the macro-trend but fail to model local transients. Our Dual-Path architecture (Blue solid) resolves this incompatibility by harmonizing global anchoring with local refinement.}
    \label{fig:motivation}
\end{figure*}

\begin{itemize}
    \item \textbf{Spectral Aliasing in Down-sampling (Figure \ref{fig:motivation}a):} To capture hierarchical dependencies, existing frameworks typically employ average pooling. However, from a signal processing perspective, pooling acts as a non-ideal low-pass filter, causing spectral aliasing, which irreversibly mixes high-frequency fluctuations into the trend, leading to information loss. To preserve spectral integrity, a mathematically orthogonal decomposition is required.

    \item \textbf{Trend-Detail Incompatibility (Figure \ref{fig:motivation}b):} Real-world time series emerge from a combination of deterministic global trends and stochastic local fluctuations. This inherent heterogeneity presents a fundamental challenge to single-stream architectures. As shown in (b), linear model successfully anchors the trend but is rigid to capture rapid transients (Missed Peak). A reasonable forecasting framework should explicitly decouple above two components.
\end{itemize}

Therefore, we bridge signal processing theory and deep representation learning by proposing DPWMixer (Dual-Path Wavelet Mixer), a novel architecture for modeling multi-scale signals with linear computational complexity.  Specifically, we replace the lossy pooling with Lossless Haar Wavelet Pyramid \cite{1198399}. Different from pooling, DWT uses orthogonal basis functions to decompose signals into approximation coefficients and detail coefficients, and DWT satisfies Parseval’s theorem (energy conservation). This prevents information loss by separating high-frequency components from trend signals. Further, to resolve conflict between trend and detail modeling, we design a Dual-Path Trend Mixer. Specifically, the module processes decomposed components into two parallel paths: a Global Linear Path anchors the prediction to macro-trend while a Local Evolution Path (implemented by a lightweight Patch-Mixer) models micro-dynamic contexts. Finally, an Adaptive Multi-Scale Fusion mechanism is implemented to weigh predictions from different scales given inputs’ characteristics. The main contributions of this paper can be summarized as follows.
\begin{itemize}
\item We formulate the Haar Wavelet Pyramid as a rigorous solution to aliasing problem in multi-scale forecasting. Both theoretical analysis and algorithm design demonstrate that, compared with pooling, Haar Wavelet Pyramid provides a better inductive bias to disentangle trends and details without information loss.
\item We design Dual-Path Trend Mixer as a hybrid layer to explicitly resolve the conflict between modeling stable global trend and flexible local variation. The design borrows the robustness of linear model and the expressiveness of MLPs.
\item We implement Adaptive Multi-Scale Fusion mechanism to weigh predictions from different resolutions adaptively, which allows the model to specialize for multivariate heterogeneity.
\item Extensive evaluations on eight benchmark datasets demonstrate that DPWMixer consistently outperforms state-of-the-art baselines, including iTransformer and TimeMixer. Furthermore, our model maintains $\mathcal{O}(L)$ time  complexity, ensuring scalability for high-throughput forecasting tasks.
\end{itemize}

The remainder of this paper is organized as follows: Section \ref{rel} reviews related work. Section \ref{met} details the DPWMixer methodology. Section \ref{exp} presents experimental results, and Section \ref{con} concludes the paper.

\section{Related Work} \label{rel}

\subsection{Transformer-based Models for LTSF}
Transformers\cite{vaswani2017attention} model the global dependence with self-attention mechanism. Its canonical quadratic complexity $\mathcal{O}(L^2)$ brings great challenges to LTSF.
Many efficient variants have been proposed to alleviate this issue. Informer\cite{zhou2021informer} proposes a ProbSparse attention mechanism based on KL-divergence criterion to select dominant queries, and the complexity is lowered to $\mathcal{O}(L \log L)$. Autoformer\cite{wu2021autoformer} abandons the point-wise attention and proposes an Auto-Correlation mechanism, which discovers the similarities between sub-series based on periodicity of series. FEDformer\cite{zhou2022fedformer} explores in the Frequency Domain, and uses Fourier and Wavelet transform to select a subset of frequency components for linear complexity interaction. Pyraformer\cite{DBLP:conf/iclr/LiuYLLLLD22} constructs a pyramidal attention graph to capture multi-resolution temporal dependencies.

Recently, researchers focus on input representation instead of model complexity. PatchTST\cite{nie2022time} divides time series into sub-series patches and uses them as input tokens. This design not only keeps the advantage of preserving local semantic information, but also greatly reduces the number of tokens, which in turn extends the look-back window. Besides, PatchTST discusses the advantages of Channel Independence, where each variate is forecasted independently. While Crossformer\cite{zhang2023crossformer} and iTransformer\cite{liu2023itransformer} argue that cross-variate dependency should be considered. Particularly, iTransformer inverts the dimensions, where the whole time series of each variate is embedded into one token and applies attention over variates to capture the multivariate correlations. Transformer based models are prone to overfitting because of the sparse semantic density of time series compared with NLP. Besides, they may also be weak in learning trend information without any decomposition modules.

\subsection{MLP and Linear-based Models}

Challenging the dominance of Transformers, DLinear\cite{zeng2023transformers} decomposes time series into trend and seasonal components and models them with only one linear layer. The simple design shows that what matters for LTSF is that the temporal order of inputs should be preserved and more complex attention might be useless or even harmful due to the permutation invariance. RLinear\cite{li2023revisitinglongtermtimeseries} further studies the influence of input Reversibility and Normalization.

Among MLPs, TiDE\cite{das2023longterm} proposes a dense encoder-decoder model that integrates exogenous variables and covariates through simple MLP layers and obtains competitive results with high throughput. TSMixer\cite{DBLP:conf/kdd/EkambaramJNSK23} extends MLP-Mixer in computer vision to time series and mixes along both time and feature dimensions to model intra- and inter-variate dependencies. These models are more efficient and stable, their receptive fields are rigid. Linear models posit a relatively fixed relationship between history and future, which is unable to model the highly complex nonlinear evolutionary dynamics, especially for time series with rapidly changing distributions (distribution shift). Moreover, simple linear mappings would be a kind of low-pass filter that may smear out important high-frequency anomalies.

\subsection{Multi-Scale and Frequency Domain Modeling}
Real-world time series exhibit intricate patterns at various granularities. Multi-scale modeling aims to capture these hierarchical dependencies. SCINet\cite{liu2022scinet} proposes a recursive down-sampling-convolve-interact architecture to learn multi-resolution representations, enhancing predictability. TimesNet\cite{DBLP:conf/iclr/WuHLZ0L23} transforms 1D series into 2D tensors to utilize convolutions. MSTF \cite{DBLP:journals/tjs/ZhouJLCZ25} proposed a multi-scale temporal fusion model using time reverse blocks and dynamic combination reconstruction. However, MSTF relies on average pooling for down-sampling, which inherently leads to information loss and aliasing of high-frequency signals according to the Nyquist-Shannon sampling theorem. 
In the frequency domain, ScaleMixNet \cite{DBLP:journals/tjs/ZhaoLZ25} proposes an adaptive multi-scale time-frequency fusion network using FFT and hybrid loss functions. While effective, Fourier-based methods provide a global view of frequencies and struggle to capture local transient changes typical in non-stationary time series.

A recent representative method TimeMixer\cite{wang2024timemixer} uses past-decomposable-mixing architecture, and average pooling to construct a multi-scale input pyramid. Although the receptive field can be enlarged by this design, since the average pooling is in fact a kind of non-invertible low-pass filter according to the Nyquist-Shannon sampling theorem, inevitable aliasing would be induced and high-frequency information (transients) necessary for forecasting volatile series will be randomly lost.

In frequency domain, FiLM\cite{zhou2022film} and ETSformer\cite{woo2022etsformer} use Legendre projections and exponential smoothing respectively to remember historical information. However, Fourier based methods usually offer global view on frequencies, while historical information of non-stationary data should be time-frequency localized. In contrast, our method employs Orthogonal Haar Wavelet Transforms in dual-path. Unlike pooling, wavelets can provide an orthogonal decomposition into approximation and detail coefficients. Unlike Fourier, wavelets could provide localization in both time and frequency simultaneously, which helps our model to capture trend in evolution and transient anomalies at the same time.

\section{Methodology} \label{met}

\subsection{Problem Formulation and Framework Overview}

Let $\mathcal{X} = \{\mathbf{x}_1, \dots, \mathbf{x}_L\} \in \mathbb{R}^{L \times C}$ denote the historical multivariate time series, where $L$ represents the look-back window size and $C$ denotes the number of variates. The fundamental objective of Long-Term Time Series Forecasting (LTSF) is to estimate the joint probability distribution $P(\mathcal{Y}|\mathcal{X})$ and predict the future sequence $\mathcal{Y} = \{\mathbf{x}_{L+1}, \dots, \mathbf{x}_{L+T}\} \in \mathbb{R}^{T \times C}$ over a forecast horizon $T$. We aim to approximate the optimal mapping function $\mathcal{F}_{\theta}: \mathbb{R}^{L \times C} \mapsto \mathbb{R}^{T \times C}$, parameterized by $\theta$, that minimizes the $L_2$ norm distance (Mean Squared Error) between the ground truth $\mathcal{Y}$ and the prediction $\hat{\mathcal{Y}}$.

\begin{figure*}[t]
\centering
\includegraphics[width=1\textwidth]{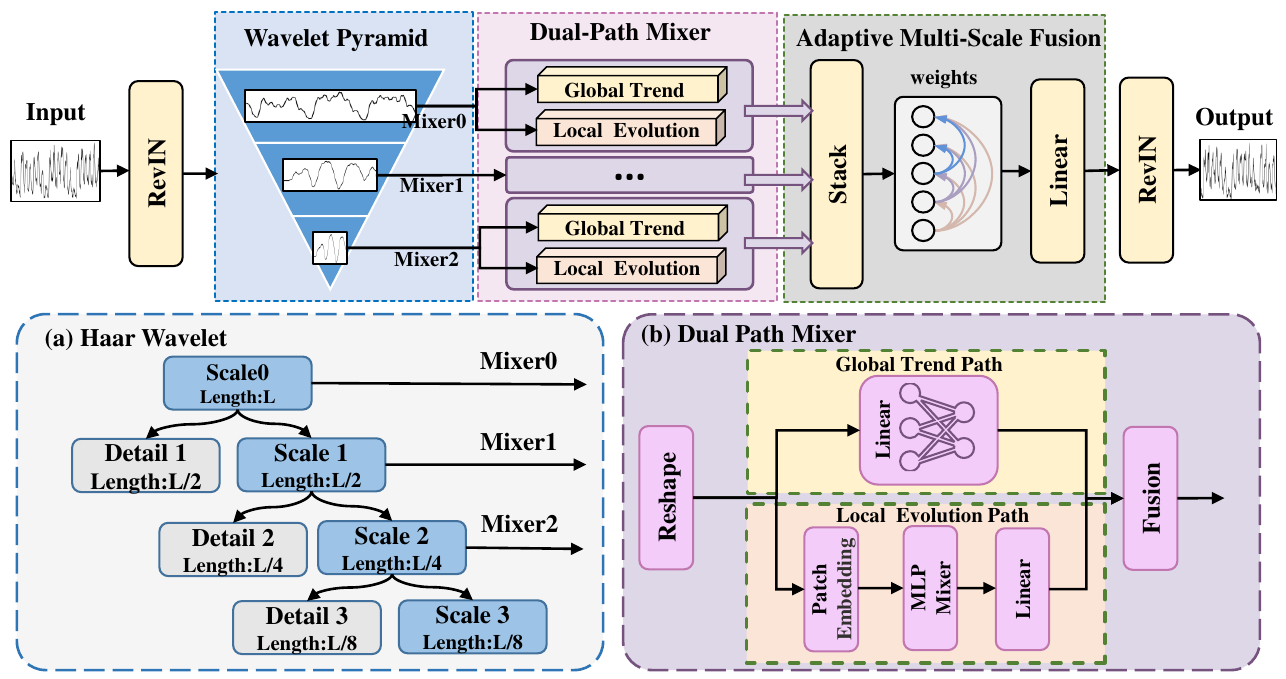}
\caption{The overall architecture of DPWMixer. Top: The global pipeline showing orthogonal multi-scale decomposition and adaptive fusion. (a) Haar Wavelet Pyramid: The input is orthogonally decomposed into Approximation ($\mathbf{X}$) and Detail ($\mathbf{H}$) coefficients, preventing aliasing compared to average pooling. (b) Dual-Path Trend Mixer: A hybrid block unifying a Global Trend Path for rigid trends and a Local Evolution Path for flexible dynamics. Outputs are fused via learnable gates.}
\label{fig:main_arch}
\end{figure*}

To bridge these gaps, our proposed DPWMixer incorporates the following two innovations: (1) Orthogonal Multi-Resolution Decomposition that recursively splits normalized series by Lossless Haar Wavelet Pyramid to preserve spectral information resulted in down-sampling aliasing; (2) Dual-Path Representation Learning that models deterministic global trend and stochastic local semantics via linear path and MLP-based path respectively at each extracted multi-scale representation. As illustrated in Figure \ref{fig:main_arch}, our method consists of three stages: orthogonal multi-resolution decomposition, which recursively splits normalized series into multi-scale representations via a Lossless Haar Wavelet Pyramid to preserve spectral information affected by down-sampling aliasing; dual-path representation learning, where at each resolution scale, a Dual-Path Trend Mixer models deterministic global trends and stochastic local semantics through a linear path and an MLP-based path, respectively; and adaptive spectral aggregation, which integrates multi-scale forecasts into the final prediction via an Adaptive Multi-Resolution Fusion module that adaptively aggregates different frequency bands weighted by channel-wise characteristics.

To synthesize the proposed framework, Algorithm \ref{alg:dpwmixer} outlines the holistic forward propagation pipeline. This procedure systematically orchestrates the lossless wavelet decomposition and dual-path mixing, ensuring that multi-scale features are extracted from physically disentangled frequency bands before adaptive integration.
\begin{algorithm}[t]
\caption{DPWMixer Training Procedure}
\label{alg:dpwmixer}
\begin{algorithmic}[1] 
    \Require Historical time series $\mathbf{X} \in \mathbb{R}^{B \times L \times C}$, Prediction horizon $T$, Number of scales $N$.
    \Ensure Predicted future series $\hat{\mathbf{Y}} \in \mathbb{R}^{B \times T \times C}$.

    \State \textbf{1. Distribution Alignment (RevIN)}
    \State $\mu, \sigma \leftarrow \text{Statistics}(\mathbf{X})$
    \State $\mathbf{X}' \leftarrow \text{Normalize}(\mathbf{X}, \mu, \sigma)$ \Comment{Instance normalization}

    \State \textbf{2. Lossless Wavelet Pyramid Construction}
    \State Initialize scale list $\mathcal{S} \leftarrow [\mathbf{X}']$
    \For{$j = 0$ to $N-1$}
        \State $\mathbf{X}^{(j)} \leftarrow \mathcal{S}[j]$
        \State $\mathbf{X}_{low}, \mathbf{X}_{high} \leftarrow \text{HaarDWT}(\mathbf{X}^{(j)})$ \Comment{Orthogonal decomposition}
        \State $\mathcal{S}.\text{append}(\mathbf{X}_{low})$ \Comment{Down-sampled approximation}
    \EndFor

    \State \textbf{3. Multi-Scale Dual-Path Mixing}
    \State Initialize prediction list $\mathcal{P} \leftarrow []$
    \For{$j = 0$ to $N$}
        \State $\mathbf{X}^{(j)} \leftarrow \mathcal{S}[j]$ \Comment{Input at scale $j$, Length $L/2^j$}
        
        \State \textit{// Path A: Global Trend Patch}
        \State $\mathbf{H}_{global} \leftarrow \mathbf{W}_{lin}^{(j)} \mathbf{X}^{(j)}$
        
        \State \textit{// Path B: Local Evolution Patch}
        \State $\mathbf{Z}_{patch} \leftarrow \text{PatchEmbedding}(\mathbf{X}^{(j)})$
        \State $\mathbf{Z}_{mix} \leftarrow \text{MLP-Mixer}(\mathbf{Z}_{patch})$ \Comment{Token \& Channel mixing}
        \State $\mathbf{H}_{local} \leftarrow \text{Projection}(\mathbf{Z}_{mix})$
        
        \State \textit{// Gated Fusion for Scale $j$}
        \State $\hat{\mathbf{Y}}^{(j)} \leftarrow w_g^{(j)} \mathbf{H}_{global} + w_l^{(j)} \mathbf{H}_{local}$ \Comment{Combine rigidity \& flexibility}
        \State $\mathcal{P}.\text{append}(\hat{\mathbf{Y}}^{(j)})$
    \EndFor

    \State \textbf{4. Adaptive Multi-Resolution Fusion}
    \State Compute weights $\mathbf{W} \leftarrow \text{Softmax}(\mathcal{A}_{learnable})$
    \State $\hat{\mathbf{Y}}' \leftarrow \sum_{j=0}^{N} \mathbf{W}[j] \odot \mathcal{P}[j]$ \Comment{Weighted ensemble}

    \State \textbf{5. Inverse Normalization}
    \State $\hat{\mathbf{Y}} \leftarrow \text{DeNormalize}(\hat{\mathbf{Y}}', \mu, \sigma)$

    \State \Return $\hat{\mathbf{Y}}$
\end{algorithmic}
\end{algorithm}

\subsection{Haar Wavelet Pyramid}
A limitation of existing multi-scale architectures is the adoption of average pooling for down-sampling. From the signal processing point of view, by convolution with a rectangular window function, average pooling leads to the multiplication of a Sinc function in the frequency domain. Since the Sinc function decays slowly in $\mathcal{O}(1/f)$, most of its energy is at low frequencies, i.e., spectral leakage. When sampling rate is reduced, high-frequency part cannot be suppressed enough fold back into lower frequency bands, aliasing. This distortion makes down-sampled representation not an accurate approximation of the original trend, breaking the effectiveness of learning long-term dependencies.

\subsubsection{Orthogonal Decomposition Mechanism}
To overcome the spectral aliasing issue, we employ Discrete wavelet transform(DWT) \cite{sundararajan2016discrete} with Haar basis. Different from pooling which is also an approximation process similar to convolving with a window function, Haar transform uses a pair of Quadrature Mirror Filters (QMF) to decompose the signal space $V_j$ into two mutually orthogonal subspaces: the approximation subspace $V_{j+1}$ and the detail subspace $W_{j+1}$. The orthogonality condition guarantees that the energy of the signal is conserved (Parseval’s Theorem) and the information is losslessly split.

We employ Haar wavelet over continuous bases (e.g., Daubechies or Symlets) for two reasons. First, from the signal processing point of view, the step-function nature of Haar basis renders it better at capturing sudden changes and transients (e.g., traffic peaks or sensor faults) without introducing the ringing artifacts produced by higher order wavelets. Second, in HPC, Haar transform provides the lowest possible computational overhead among all wavelets. Given its filter length is only 2, it incurs the least floating point operations and memory access cost. Therefore, it strictly preserves the linear complexity $\mathcal{O}(L)$ of our whole framework.

We construct the pyramid recursively. Let $\mathbf{X}^{(0)} = \mathbf{X}'$ be the normalized input. For each decomposition level $j$, the approximation coefficients $\mathbf{X}^{(j+1)}$ and detail coefficients $\mathbf{D}^{(j+1)}$ are computed via 1D convolution with a specific stride:

\begin{equation}
    \mathbf{X}^{(j+1)} = \left(\mathbf{X}^{(j)} \ast \mathbf{k}_{low}\right) \downarrow 2
\end{equation}
\begin{equation}
    \mathbf{D}^{(j+1)} = \left(\mathbf{X}^{(j)} \ast \mathbf{k}_{high}\right) \downarrow 2
\end{equation}

where $\mathbf{k}_{low} = [1/\sqrt{2}, 1/\sqrt{2}]$ and $\mathbf{k}_{high} = [1/\sqrt{2}, -1/\sqrt{2}]$ are fixed Haar filters. 

The notation $\downarrow 2$ represents the down-sampling layer which is implemented as a strided convolution (stride=2, padding=0). Since $L^{(j)}$ may not be divisible by 2 (i.e., non-power-of-2 length), there might be some troubles in handling arbitrary sequence lengths. To avoid wasting information due to boundary effects, we explicitly apply symmetric padding on the tail of sequence before convolution. The down-sampled outputs are always accurately calibrated without cutting information at boundaries. Finally, we transmit the approximation coefficients $\mathbf{X}^{(j+1)}$ to the upper layers while suppressing the high-frequency noises without introducing aliasing artifacts.

\subsection{Dual-Path Trend Mixer}
Time series data inherently exhibits a kinematic duality: a deterministic macroscopic trend (e.g., seasonality, monotonic growth) superimposed with stochastic microscopic fluctuations. Single-stream models are challenged to model these two components in a disentangled manner. We design the Dual-Path Trend Mixer at each scale $j$, including Global Trend Modeling Path and Local Evolution Path.

\subsubsection{Global Trend Path}
Such path is intended to model low-frequency, global trajectory of time series. Linear models have shown to be more robust when extrapolating monotonic trends compared to deep neural networks which are prone to fit local noises. Thus, we apply channel-independent linear projection to directly map entire historical sequence at scale $j$ onto prediction horizon:
\begin{equation}
    \mathbf{H}_{global}^{(j)} = \mathbf{W}_{lin}^{(j)} \cdot \text{Flatten}(\mathbf{X}^{(j)}) + \mathbf{b}_{lin}^{(j)}
\end{equation}
where $\mathbf{W}_{lin}^{(j)} \in \mathbb{R}^{T \times (L^{(j)} \cdot C)}$ is the weight matrix sharing parameters across channels. This component serves as a global constraint, ensuring that the forecasted trajectory adheres to the global inertia of the historical data.

\subsubsection{Local Evolution Path}
Although the linear path provides the global direction, it is still unable to capture the second-order non-linear dynamics and local semantic variation. This path designs a Patching + MLP-Mixer module to capture local evolution path.
\begin{itemize}
    \item Patching Operation: The input sequence $\mathbf{X}^{(j)} \in \mathbb{R}^{L^{(j)} \times C}$ is segmented into $N_p$ non-overlapping patches of length $P$. This transforms the 2D temporal variates into a 3D tensor $\mathcal{Z}_{patch} \in \mathbb{R}^{N_p \times P \times C}$, preserving the local temporal context within each patch.
    \item MLP-Mixer Backbone: We project the patches into a hidden dimension $D$ and process them through a stack of Mixer layers. Each layer consists of two distinct MLP blocks:
    \begin{equation}
        \mathbf{U} = \mathbf{Z} + \text{MLP}_{token}(\text{LayerNorm}(\mathbf{Z}))
    \end{equation}
    \begin{equation}
        \mathbf{Z}_{out} = \mathbf{U} + \text{MLP}_{channel}(\text{LayerNorm}(\mathbf{U}))
    \end{equation}
    The Token-Mixing MLP captures temporal dependencies across different time segments, while the Channel-Mixing MLP extracts feature correlations within each patch.
    \item {Projection}: The output features are flattened and projected to the target horizon $T$ via a linear head, yielding the local component $\mathbf{H}_{local}^{(j)}$.
\end{itemize}

The prediction at scale $j$  is synthesized by fusing the deterministic and stochastic components via learnable scalar gates $w_g$ and $w_l$:
\begin{equation}
    \hat{\mathbf{Y}}^{(j)} = w_g \cdot \mathbf{H}_{global}^{(j)} + w_l \cdot \mathbf{H}_{local}^{(j)}
\end{equation}
This formulation can be interpreted as a residual learning framework, where the MLP path learns the non-linear residuals that the linear path cannot approximate.

\subsection{Adaptive Multi-Scale Fusion}
A critical observation often overlooked in prior work is that the optimal scales are heterogeneous across different variables: e.g., for a voltage sensor signal, the signal may be mostly contaminated by high-frequency noises and hence needs to be predicted based on relatively coarse scales ($j=2,3$); while for a traffic flow signal containing sharp and informative peaks, the model should focus on finer scales ($j=0,1$).
In this work, we explore the channel-aware adaptive fusion mechanism. Suppose $\mathcal{A} \in \mathbb{R}^{(N+1) \times C}$ is a learnable attention parameter matrix, then the fusion weights $\mathbf{W}_{fusion}$ will be rescaled by Softmax over scale dimension:
\begin{equation}
    \mathbf{W}_{fusion}[k, c] = \frac{\exp(\mathcal{A}_{k,c})}{\sum_{n=0}^{N} \exp(\mathcal{A}_{n,c})}
\end{equation}
where $k$ denotes the scale index and $c$ denotes the channel index. The final prediction $\hat{\mathbf{Y}}$ is computed as the weighted sum of forecasts from all scales:
\begin{equation}
    \hat{\mathbf{Y}}_{norm} = \sum_{j=0}^{N} \mathbf{W}_{fusion}[j] \odot \hat{\mathbf{Y}}^{(j)}
\end{equation}
where $\odot$ denotes element-wise multiplication with broadcasting. With such design, our model is able to automatically focus on the best frequency band for each variable and hence acts as a learnable spectral filter.

\subsection{Complexity Analysis}
We present a formal complexity analysis to show that DPWMixer remains linear despite its multi-scale design. Suppose $L$ is the input length and $T$ is the prediction horizon.
\begin{itemize}
\item Scale 0: The complexity is dominated by Mixer ($\mathcal{O}(L \cdot D^2)$ given fixed patch size) and Linear path ($\mathcal{O}(L \cdot T)$).
\item Scale j: The sequence length becomes $L/2^j$. Therefore, the complexity of $j$-th mixer is $\mathcal{O}(\frac{L}{2^j} \cdot D^2)$.
\end{itemize}
The total complexity is the summation of geometric series:
\begin{equation}
\mathcal{C}_{total} = \sum_{j=0}^{N} \mathcal{C}_{total}(\frac{L}{2^j}) \approx \mathcal{C}_{total}(L) \cdot (1 + \frac{1}{2} + \frac{1}{4} + \dots) \le 2 \cdot \mathcal{C}_{total}(L)
\end{equation}
Therefore, the derivation shows that when we add the multi-scale pyramid, the computational cost only increases by a constant (at most 2 times) compared with a single-scale model. Thus, DPWMixer is proved to have $\mathcal{O}(L)$ complexity. While the complexity of Transformer models is $\mathcal{O}(L^2)$ or $\mathcal{O}(L \log L)$, DPWMixer is more efficient, especially when the look-back window is longer.

\section{Experiments} \label{exp}

To comprehensively evaluate the efficacy of DPWMixer ,  we perform extensive experiments on eight real-world benchmarks. Our evaluation is not limited to simple comparisons. We empirically validate our theoretical claims on orthogonal decomposition, dual-path modeling, and adaptive fusion, respectively.

\subsection{Experimental Setup}

\subsubsection{Datasets}
We employ eight multivariate time series benchmarks from different domains. These datasets are challenging in different aspects in terms of periodicity, trend, and noise. We summarize statistical information in Table \ref{tab:dataset_stats}.
\begin{itemize}

\item \textit{ETT (Electricity Transformer Temperature)}  Comprising four subsets: ETTh1, ETTh2 (hourly) and ETTm1, ETTm2 (15-minutely). This dataset records the load and oil temperature of electricity transformer. This dataset has strong long-term seasonal periodicity but has high local volatility due to the change of load.
\item \textit{Electricity}  This dataset records the hourly electricity consumption of $321$ clients. Compared with ETT, the consumption behavior of each client has shifted to a different style over time. This set tests the robustness of models against distribution shifts.
\item \textit{Weather}  This dataset records $21$ different meteorological factors (e.g., air temperature, humidity) at every $10$ minutes. Weather is known to be a very chaotic factor. It follows thermodynamic rules, local changes are stochastic and less trended compared with energy.
\item \textit{Exchange} This dataset records the daily exchange rate of $8$ countries. This dataset is dominated by aleatoric uncertainty and has no obvious seasonal periodicity. This dataset is a stress test for financial trend.
\item \textit{Traffic}  This dataset records the road occupancy rate from $862$ sensors over San Francisco Bay Area freeways. This is a typical high-frequency dataset. Occupancy has sharp and non-linear peaks (rush hours) and is obviously periodic every day and every week. 
\end{itemize}

\begin{table}[h!]
\centering
\caption{STATISTICS OF DATASETS IN THE EXPERIMENT}
\label{tab:dataset_stats}
\begin{tabular}{lcccc}
\toprule
\textbf{Dataset} & \textbf{\#Variates} & \textbf{Timesteps} & \textbf{Frequency} & \textbf{Domain} \\
\midrule
ETTh1       & 7   & 17,420 & 1-hour  & Energy \\
ETTh2       & 7   & 17,420 & 1-hour  & Energy \\
ETTm1       & 7   & 69,680 & 15-min  & Energy \\
ETTm2       & 7   & 69,680 & 15-min  & Energy \\
Electricity & 321 & 26,304 & 1-hour  & Energy \\
Weather     & 21  & 52,696 & 10-min  & Weather \\
Exchange    & 8   & 7,588  & Daily   & Economy \\
Traffic     & 862 & 17,544 & 1-hour  & Transportation \\
\bottomrule
\end{tabular}
\end{table}

\subsubsection{Baselines}
We benchmark DPWMixer against 9 SOTA models, categorizing baselines by their core architectures to ensure a fair evaluation.
\textbf{Transformer-based Models}:
\begin{itemize}
\item \textit{iTransformer} \cite{liu2023itransformer} Reverse the mainstream Transformer by embedding the whole time series of each variate into one token. Then apply attentions over tokens among different variates to explicitly capture the multivariate correlation. Finally, it achieves the state-of-the-art performance.
\item \textit{PatchTST} \cite{nie2022time} Divide the time series into sub-series patches to keep the local semantic information and reduce the computational cost. Then, it adopts a channel-independent strategy to allow a shared Transformer backbone to learn the generalizable temporal pattern from different variables.
\item \textit{Crossformer} \cite{zhang2023crossformer} Employ a Two-Stage Attention (TSA) mechanism to capture the dependency of target series from other patches spanned over both time and dimension. It effectively routes the information among patches from different variates. Different from channel-independent methods which model the interaction between channels as modeling the interaction between dimensions, patches from different variates are highly correlated.
\item \textit{FEDformer} \cite{zhou2022fedformer} Adopt seasonal-trend decomposition and frequency domain analysis. It uses Fourier/Wavelet transform to extract k significant frequency components and then adopts the learned components as inputs of linearformer. Therefore, the seasonal-trend decomposition makes FEDformer linear complexity attention while capturing the global periodic pattern effectively.
\item \textit{TimeXer} \cite{wang2024timexer} Adopts the patch-wise inverted architecture to leverage exogenous variables effectively. It utilizes an Endogenous-Exogenous Attention mechanism that adaptively aligns and aggregates information from external series to the target series, ensuring that only beneficial correlations are captured while filtering out irrelevant noise.

\end{itemize}
\textbf{MLP/Linear-based Models}:
\begin{itemize}
\item \textit{TimeMixer} \cite{wang2024timemixer} Design a Past-Decomposable-Mixing (PDM) architecture. It uses average pooling to down-sample the series into multi-scale pyramid and then apply MLP-based mixing operations on the series at different temporal resolutions. It may lead to aliasing since it mixes information from different temporal resolutions.
\item \textit{DLinear} \cite{zeng2023transformers} Decompose time series into trend and seasonal components using moving average kernels. It models each component with a simple single-layer linear mapping. It shows that preserving the temporal order is more important than a complex network design on each time series.
\item \textit{FITS} \cite{DBLP:conf/iclr/XuZ024} FITS uses frequency domain interpolation to achieve state-of-the-art performance in time series forecasting and anomaly detection tasks. This lightweight model is parameter-efficient compared with complex Transformers and more efficient, universal, and adaptive to different variable input lengths and analytical tasks.
\end{itemize}
\textbf{CNN-based Models}:
\begin{itemize}
\item \textit{TimesNet} \cite{DBLP:conf/iclr/WuHLZ0L23}: Rearranges 1D time series into 2D tensors according to multi-periodicity analysis. A parameter-efficient Inception block is used to model intra-period and inter-period variations in a complementary way, which efficiently extends 2D visual backbones to TS forecasting.
\end{itemize}

\subsubsection{Implementation Details}

All experiments are conducted based on PyTorch in a cluster with NVIDIA RTX 4090 GPUs.
Following the standard protocol of Long-Term Time Series Forecasting LTSF, we also test the forecasting performance on another three long-term prediction horizons $T \in \{96, 192, 336, 720\}$ to cover the forecasting range from short to long.

For reproducibility and fairness, we follow the following strict protocols.
Data Splitting. Following the standard benchmarks in Informer \cite{zhou2021informer}, we follow the 7:1:2 ratio to split all datasets into training set, validation set and testing set. For Traffic dataset, we follow 6:2:2 ratio since it could reflect the spatial-temporal dynamics more complicatedly. $L = 96$.

We adopt ADAM optimizer with cosine annealing learning rate scheduler. The learning rate will be selected from  $\{10^{-4}, 5 \times 10^{-4}, 10^{-3}\}$, through grid search, and the batch size will be chosen from ${16,32,64}$ according to the size of the corresponding dataset. To avoid over-fitting, we adopt an Early Stopping mechanism on top of the validation loss, which is implemented with patience of 5 epochs. The maximum epoch is limited in 10. The number of scales $N$ in wavelet decomposition is set to 3, which is a rational design to balance the expansion of the receptive field and the preservation of resolution. We unify the patch length P in 16 and set the latent hidden dimension D in 128. All results reported in the paper are the mean of 3 runs with different random seeds.

Evaluation Metrics. We adopt two standard quantitative metrics: Mean Squared Error (MSE) and Mean Absolute Error (MAE). MSE is the mean of the squared error between the predicted value and real value. MSE will be strictly punished on the large error to identify large deviation in volatile series. While MAE is a quite robust way to evaluate the general performance of the forecast on the basis of absolute error. For both of MSE and MAE, the smaller value means the better performance.

\begin{sidewaystable*}[p]
\centering
\renewcommand{\arraystretch}{1.4}
\setlength{\tabcolsep}{1.1pt} 
\caption{Long-term forecasting performance comparison with an input length of $L=96$ for prediction horizons $T \in \{96, 192, 336, 720\}$. Best and second-best scores are marked in \textbf{bold} and with an \underline{underline}, respectively. 'Avg' denotes the average performance.}
\resizebox{\linewidth}{!}{%
 \begin{tabular}{l l *{20}{w{r}{3em}}} 
\toprule
\multicolumn{2}{l}{\textbf{Models}} & \multicolumn{2}{c}{\textbf{DPWMixer}} & \multicolumn{2}{c}{\textbf{iTransformer}}  &\multicolumn{2}{c} {\textbf{TimeMixer}}& \multicolumn{2}{c}{\textbf{TimeXer}} & \multicolumn{2}{c}{\textbf{FITS}} & \multicolumn{2}{c}{\textbf{PatchTST}} & \multicolumn{2}{c}{\textbf{Crossformer}} & \multicolumn{2}{c}{\textbf{TimesNet}} & \multicolumn{2}{c}{\textbf{DLinear}}  & \multicolumn{2}{c}{\textbf{FEDformer}} \\
\multicolumn{2}{l}{} & \multicolumn{2}{c}{\textbf{(Ours)}} & \multicolumn{2}{c}{\textbf{2024}} & \multicolumn{2}{c}{\textbf{2024}}  & \multicolumn{2}{c}{\textbf{2024}} & \multicolumn{2}{c}{\textbf{2024}}& \multicolumn{2}{c}{\textbf{2023}}  & \multicolumn{2}{c}{\textbf{2023}} & \multicolumn{2}{c}{\textbf{2023}} & \multicolumn{2}{c}{\textbf{2023}} & \multicolumn{2}{c}{\textbf{2022}}  \\
\cmidrule(lr){3-4} \cmidrule(lr){5-6} \cmidrule(lr){7-8} \cmidrule(lr){9-10} \cmidrule(lr){11-12} \cmidrule(lr){13-14} \cmidrule(lr){15-16} \cmidrule(lr){17-18} \cmidrule(lr){19-20} \cmidrule(lr){21-22} 
\textbf{Metric} & & \textbf{MSE} & \textbf{MAE} & \textbf{MSE} & \textbf{MAE} & \textbf{MSE} & \textbf{MAE} & \textbf{MSE} & \textbf{MAE} & \textbf{MSE} & \textbf{MAE} & \textbf{MSE} & \textbf{MAE} & \textbf{MSE} & \textbf{MAE} & \textbf{MSE} & \textbf{MAE} & \textbf{MSE} & \textbf{MAE} & \textbf{MSE} & \textbf{MAE} \\
\midrule
\multirow{5}{*}{{ETTh1}} & 96    & {\textbf{0.373}} & {\textbf{0.391}} & 0.386 & 0.405 & \underline{0.375} & \underline{0.4} & 0.382 & 0.403 & 0.450  & 0.442 & 0.46  & 0.447 & 0.423 & 0.448 & 0.384 & 0.402 & 0.407 & 0.412 & 0.395 & 0.424 \\
          & 192   & {\textbf{0.426}} & {\textbf{0.419}} & 0.441 & 0.512 & \underline{0.429} & \underline{0.421} & 0.429 & 0.453 & 0.522 & 0.481 & 0.477 & 0.429 & 0.471 & 0.474 & 0.436 & 0.446 & 0.441 & 0.411 & 0.469 & 0.470 \\
          & 336   & {\textbf{0.462}} & {\textbf{0.441}} & 0.487 & 0.458 & 0.484 & 0.458 & \underline{0.468} &\underline{ 0.448} & 0.553 & 0.501 & 0.546 & 0.496 & 0.496 & 0.470  & 0.491 & 0.491 & 0.469 & 0.489 & 0.547 & 0.495 \\
          & 720   & \underline{0.491} & \underline{0.474} & 0.503 & 0.491 & 0.498 & 0.482 & {\textbf{0.469}} & {\textbf{0.461}} & 0.545 & 0.517 & 0.544 & 0.517 & 0.653 & 0.621 & 0.521 & 0.500   & 0.513 & 0.510  & 0.598 & 0.544 \\
\cmidrule(lr){2-22}
     &Avg & {\textbf{0.438}} & {\textbf{0.431}} & 0.454 & 0.447 & 0.440  & 0.438 & \underline{0.437} & \underline{0.437} & 0.517 & 0.485 & 0.516 & 0.484 & 0.529 & 0.522 & 0.458 & 0.450  & 0.461 & 0.457 & 0.498 & 0.484 \\
\midrule
\multirow{5}{*}{{ETTh2}} & 96    & {\textbf{0.282}} & {\textbf{0.326}} & \underline{0.286} & \underline{0.338} & 0.289 & 0.342 & 0.308 & 0.355 & 0.314 & 0.359 & 0.745 & 0.584 & 0.400   & 0.440  & 0.340  & 0.374 & 0.340  & 0.394 & 0.358 & 0.397 \\
          & 192   & {\textbf{0.361}} & {\textbf{0.387}} & 0.380  & 0.400   & 0.378 & 0.397 & \underline{0.363} & \underline{0.389} & 0.406 & 0.414 & 0.793 & 0.585 & 0.877 & 0.656 & 0.402 & 0.452 & 0.419 & 0.479 & 0.414 & 0.439 \\
          & 336   & \underline{0.409} & \underline{0.421} & 0.428 & 0.432 & {\textbf{0.386}} & {\textbf{0.414}} & 0.414 & \underline{0.423} & 0.446 & 0.447 & 0.927 & 0.643 & 1.043 & 0.731 & 0.452 & 0.482 & 0.591 & 0.541 & 0.496 & 0.487 \\
          & 720   & \underline{0.423} & {\textbf{0.426}} & 0.427 & 0.445 & {\textbf{0.412}} & 0.434 & \underline{0.414} & \underline{0.432} & 0.475 & 0.471 & 1.043 & 0.636 & 1.104 & 0.763 & 0.462 & 0.468 & 0.661 & 0.661 & 0.463 & 0.474 \\
\cmidrule(lr){2-22}
          &Avg & \underline{0.368} & {\textbf{0.39}} & 0.383 & 0.407 & {\textbf{0.364}} & 0.398 & 0.383 & \underline{0.396} & 0.410  & 0.422 & 0.878 & 0.612 & 0.841 & 0.642 & 0.414 & 0.427 & 0.563 & 0.519 & 0.437 & 0.449 \\
\midrule
\multirow{5}{*}{ETTm1} & 96    & {\textbf{0.316}} & {\textbf{0.346}} & 0.334 & 0.368 & 0.320  & 0.357 & \underline{0.318} & \underline{0.356} & 0.350  & 0.370  & 0.352 & 0.374 & 0.404 & 0.426 & 0.338 & 0.375 & 0.346 & 0.374 & 0.379 & 0.419 \\
          & 192   & {\textbf{0.356}} & {\textbf{0.373}} & 0.404 & 0.393 & \underline{0.361} & 0.393 & 0.362 & \underline{0.383} & 0.392 & 0.393 & 0.387 & 0.404 & 0.450  & 0.451 & 0.374 & 0.387 & 0.381 & 0.391 & 0.389 & 0.387 \\
          & 336   & {\textbf{0.378}} & {\textbf{0.401}} & 0.426 & 0.420  & \underline{0.39} & \underline{0.404} & 0.395 & 0.407 & 0.424 & 0.414 & 0.421 & 0.414 & 0.532 & 0.515 & 0.410  & 0.411 & 0.415 & 0.415 & 0.445 & 0.459 \\
          & 720   & 0.452 & {\textbf{0.437}} & 0.491 & 0.459 & 0.454 & \underline{0.441} & {\textbf{0.441}} & 0.441 & 0.484 & \underline{0.447} & 0.462 & 0.449 & 0.666 & 0.589 & 0.478 & 0.450  & 0.473 & 0.451 & 0.543 & 0.490 \\
\cmidrule(lr){2-22}
                &Avg & {\textbf{0.375}} & {\textbf{0.389}} & 0.407 & 0.410  & \underline{0.381} & 0.395 & 0.382 & \underline{0.391} & 0.412 & 0.406 & 0.406 & 0.407 & 0.513 & 0.495 & 0.400   & 0.406 & 0.404 & 0.408 & 0.448 & 0.452 \\
\midrule
\multirow{5}{*}{ETTm2}
& 96    & {\textbf{0.169}} & {\textbf{0.255}} & 0.180  & 0.264 & 0.175 & 0.258 & \underline{0.171} & \underline{0.256} & 0.184 & 0.268 & 0.183 & 0.270  & 0.287 & 0.366 & 0.187 & 0.267 & 0.193 & 0.286 & 0.203 & 0.287 \\
          & 192   & {\textbf{0.232}} & {\textbf{0.297}} & 0.250  & 0.309 & \underline{0.237} & \underline{0.299} & 0.237 & 0.299 & 0.249 & 0.307 & 0.255 & 0.314 & 0.414 & 0.492 & 0.249 & 0.309 & 0.284 & 0.361 & 0.269 & 0.328 \\
          & 336   & {\textbf{0.293}} & {\textbf{0.335}} & 0.311 & 0.348 & 0.298 & 0.340  & \underline{0.296} & \underline{0.338} & 0.309 & 0.343 & 0.309 & 0.347 & 0.597 & 0.542 & 0.321 & 0.331 & 0.382 & 0.429 & 0.325 & 0.366 \\
          & 720   & {\textbf{0.389}} & {\textbf{0.391}} & 0.407 & 0.407 & \underline{0.391} & \underline{0.392} & 0.392 & 0.394 & 0.409 & 0.398 & 0.412 & 0.404 & 1.730  & 1.042 & 0.408 & 0.403 & 0.558 & 0.525 & 0.421 & 0.415 \\
\cmidrule(lr){2-22}
&Avg & {\textbf{0.271}} & {\textbf{0.319}} & 0.288 & 0.332 & 0.275 & 0.323 & \underline{0.274} & \underline{0.322} & 0.287 & 0.329 & 0.290  & 0.334 & 0.757 & 0.610  & 0.291 & 0.333 & 0.354 & 0.402 & 0.305 & 0.349 \\
\bottomrule
\end{tabular}
}
\label{tab:main_results}
\end{sidewaystable*}

\begin{sidewaystable*}[p]
\centering
\renewcommand{\arraystretch}{1.4}
\setlength{\tabcolsep}{1.1pt} 
\addtocounter{table}{-1}
\caption{(continued)}
\resizebox{\linewidth}{!}{%
 \begin{tabular}{l l *{20}{w{r}{3em}}} 
\toprule
\multicolumn{2}{l}{\textbf{Models}} & \multicolumn{2}{c}{\textbf{DPWMixer}} & \multicolumn{2}{c}{\textbf{iTransformer}}  &\multicolumn{2}{c} {\textbf{TimeMixer}}& \multicolumn{2}{c}{\textbf{TimeXer}} & \multicolumn{2}{c}{\textbf{FITS}} & \multicolumn{2}{c}{\textbf{PatchTST}} & \multicolumn{2}{c}{\textbf{Crossformer}} & \multicolumn{2}{c}{\textbf{TimesNet}} & \multicolumn{2}{c}{\textbf{DLinear}}  & \multicolumn{2}{c}{\textbf{FEDformer}} \\
\multicolumn{2}{l}{} & \multicolumn{2}{c}{\textbf{(Ours)}} & \multicolumn{2}{c}{\textbf{2024}} & \multicolumn{2}{c}{\textbf{2024}}  & \multicolumn{2}{c}{\textbf{2024}} & \multicolumn{2}{c}{\textbf{2024}}& \multicolumn{2}{c}{\textbf{2023}}  & \multicolumn{2}{c}{\textbf{2023}} & \multicolumn{2}{c}{\textbf{2023}} & \multicolumn{2}{c}{\textbf{2023}} & \multicolumn{2}{c}{\textbf{2022}}  \\
\cmidrule(lr){3-4} \cmidrule(lr){5-6} \cmidrule(lr){7-8} \cmidrule(lr){9-10} \cmidrule(lr){11-12} \cmidrule(lr){13-14} \cmidrule(lr){15-16} \cmidrule(lr){17-18} \cmidrule(lr){19-20} \cmidrule(lr){21-22} 
\textbf{Metric} & & \textbf{MSE} & \textbf{MAE} & \textbf{MSE} & \textbf{MAE} & \textbf{MSE} & \textbf{MAE} & \textbf{MSE} & \textbf{MAE} & \textbf{MSE} & \textbf{MAE} & \textbf{MSE} & \textbf{MAE} & \textbf{MSE} & \textbf{MAE} & \textbf{MSE} & \textbf{MAE} & \textbf{MSE} & \textbf{MAE}  & \textbf{MSE} & \textbf{MAE}  \\
\midrule
\multirow{5}{*}{{Electricity}} & 96    & \underline{0.152} & \underline{0.245} & {\textbf{0.148}} & {\textbf{0.24}} & 0.153 & 0.247 & 0.182 & 0.278 & 0.203 & 0.282 & 0.190  & 0.296 & 0.219 & 0.314 & 0.168 & 0.272 & 0.210  & 0.305 & 0.169 & 0.273 \\
          & 192   & {\textbf{0.160}} & {\textbf{0.251}} & \underline{0.162} & \underline{0.253} & 0.166 & 0.256 & 0.199 & 0.263 & 0.201 & 0.283 & 0.196 & 0.304 & 0.231 & 0.322 & 0.184 & 0.298 & 0.210  & 0.305 & 0.201 & 0.315 \\
          & 336   & {\textbf{0.173}} & {\textbf{0.266}} & \underline{0.178} & \underline{0.269} & 0.185 & 0.277 & 0.193 & 0.312 & 0.215 & 0.297 & 0.217 & 0.319 & 0.246 & 0.337 & 0.198 & 0.300   & 0.223 & 0.319 & 0.200   & 0.304 \\
          & 720   & {\textbf{0.223}} & {\textbf{0.306}} & \underline{0.225} & \underline{0.310} & 0.225 & 0.317 & 0.233 & 0.312 & 0.257 & 0.330  & 0.258 & 0.352 & 0.280  & 0.363 & 0.220  & 0.320  & 0.258 & 0.350  & 0.246 & 0.355 \\
\cmidrule(lr){2-22}
& Avg & {\textbf{0.177}} & {\textbf{0.267}} & \underline{0.178} & \underline{0.271} & 0.182 & 0.274 & 0.202 & 0.290  & 0.219 & 0.298 & 0.216 & 0.318 & 0.244 & 0.334 & 0.192 & 0.304 & 0.225 & 0.319 & 0.214 & 0.327 \\
\midrule
\multirow{5}{*}{{Weather}} & 96    & \underline{0.166} & \underline{0.209} & 0.174 & 0.212 & 0.163 & 0.209 & {\textbf{0.157}} & {\textbf{0.205}} & 0.174 & 0.224 & 0.186 & 0.227 & 0.195 & 0.271 & \underline{0.172} & 0.220  & 0.195 & 0.252 & 0.217 & 0.296 \\
          & 192   & {\textbf{0.202}} & {\textbf{0.238}} & 0.221 & 0.254 & 0.208 & \underline{0.250} & \underline{0.204} & 0.247 & 0.223 & 0.264 & 0.234 & 0.265 & 0.209 & 0.277 & 0.219 & 0.261 & 0.237 & 0.282 & 0.276 & 0.336 \\
          & 336   & \underline{0.258} & {\textbf{0.275}} & 0.278 & 0.296 & {\textbf{0.251}} & \underline{0.278} & 0.261 & 0.290  & 0.28  & 0.302 & 0.284 & 0.301 & 0.273 & 0.332 & 0.280  & 0.306 & 0.282 & 0.331 & 0.339 & 0.380 \\
          & 720   & {\textbf{0.336}} & {\textbf{0.338}} & 0.358 & 0.347 & \underline{0.339} & \underline{0.341} & 0.340  & 0.341 & 0.357 & 0.351 & 0.356 & 0.349 & 0.379 & 0.401 & 0.365 & 0.359 & 0.359 & 0.345 & 0.403 & 0.428 \\
\cmidrule(lr){2-22}
& Avg & {\textbf{0.240}} & {\textbf{0.265}} & 0.258 & 0.278 & 0.250  & 0.283 & \underline{0.240} & \underline{0.271} & 0.258 & 0.285 & 0.265 & 0.285 & 0.264 & 0.32  & 0.259 & 0.287 & 0.265 & 0.315 & 0.309 & 0.360 \\
\midrule

\multirow{5}{*}{{Exchange}} & 96   & {\textbf{0.084}} & {\textbf{0.201}} & 0.086 & 0.206 & 0.090  & 0.235 & \underline{0.085} & \underline{0.204} & 0.085 & 0.236 & 0.088 & 0.205 & 0.256 & 0.367 & 0.107 & 0.234 & 0.088 & 0.218 & 0.148 & 0.278 \\
          & 192   & {\textbf{0.171}} & {\textbf{0.297}} & \underline{0.177} & \underline{0.299} & 0.187 & 0.343 & 0.181 & 0.302 & 0.193 & 0.392 & 0.176 & 0.299 & 0.470  & 0.509 & 0.226 & 0.344 & 0.176 & 0.315 & 0.271 & 0.315 \\
          & 336   & \underline{0.341} & {\textbf{0.413}} & {\textbf{0.331}} & \underline{0.417} & 0.353 & 0.473 & 0.363 & 0.435 & 0.386 & 0.503 & 0.301 & 0.397 & 1.268 & 0.883 & 0.367 & 0.448 & 0.313 & 0.427 & 0.460  & 0.427 \\
          & 720   & {\textbf{0.842}} & {\textbf{0.688}} & \underline{0.847} & \underline{0.691} & 0.934 & 0.761 & 0.930  & 0.727 & 1.023 & 0.862 & 0.901 & 0.714 & 1.767 & 1.068 & 0.964 & 0.746 & 0.839 & 0.695 & 1.195 & 0.695 \\
\cmidrule(lr){2-22}
& Avg & {\textbf{0.359}} & {\textbf{0.399}} & \underline{0.36} & \underline{0.403} & 0.391 & 0.453 & 0.389 & 0.417 & 0.421 & 0.498 & 0.367 & 0.404 & 0.940  & 0.707 & 0.416 & 0.443 & 0.354 & 0.414 & 0.519 & 0.429 \\
\midrule

\multirow{5}{*}{{Traffic}} & 96    & 0.452 & \underline{0.277} & {\textbf{0.395}} & {\textbf{0.271}} & 0.462 & 0.285 & \underline{0.428} & {\textbf{0.271}} & 0.725 & 0.484 & 0.526 & 0.347 & 0.644 & 0.429 & 0.593 & 0.321 & 0.650  & 0.396 & 0.587 & 0.366 \\
          & 192   & 0.469 & 0.291 & {\textbf{0.417}} & {\textbf{0.276}} & 0.473 & 0.296 & \underline{0.448} & \underline{0.282} & 0.737 & 0.512 & 0.522 & 0.332 & 0.665 & 0.431 & 0.617 & 0.336 & 0.598 & 0.370  & 0.604 & 0.373 \\
          & 336   & 0.486 & \underline{0.295} & {\textbf{0.433}} & 0.298 & 0.498 & 0.296 & \underline{0.473} & {\textbf{0.289}} & 0.730  & 0.495 & 0.517 & 0.334 & 0.674 & 0.420  & 0.629 & 0.336 & 0.605 & 0.373 & 0.621 & 0.383 \\
          & 720   & \underline{0.503} & 0.311 & {\textbf{0.467}} & {\textbf{0.302}} & 0.506 & 0.313 & 0.516 & \underline{0.307} & 0.752 & 0.492 & 0.552 & 0.352 & 0.683 & 0.424 & 0.640  & 0.350  & 0.645 & 0.394 & 0.626 & 0.382 \\
\cmidrule(lr){2-22}
& Avg  & 0.477 & 0.293 & {\textbf{0.428}} & {\textbf{0.282}} & 0.484 & 0.298 & \underline{0.466} & \underline{0.287} & 0.466 & 0.495 & 0.529 & 0.341 & 0.667 & 0.426 & 0.62  & 0.336 & 0.625 & 0.383 & 0.610  & 0.376 \\
\midrule
\multicolumn{2}{l}{1st Count} & \textbf{20}    & \textbf{24}    & 7     & 5     & 4     & 1     & 3     & 4     & 0     & 0     & 0     & 0     & 0     & 0     & 0     & 0     & 0     & 0     & 0     & 0 \\
\bottomrule
\end{tabular}
}
\label{tab:main_results2}
\end{sidewaystable*}

\subsection{Main Results}
As shown in Table \ref{tab:main_results2}, DPWMixer demonstrates state-of-the-art performance in a variety of multivariate long-term time series forecasting tasks. Compared with other strong baselines, i.e., current SOTA Transformer-based method iTransformer and multi-scale MLP method TimeMixer, our method obtains the lowest MSE and MAE in the majority of cases.

Specifically, DPWMixer ranks first in 44 out of 64 experimental settings (aggregating across 8 datasets and 4 prediction horizons), showcasing its robust generalization capability. On datasets characterized by complex periodic patterns and rapid fluctuations, such as Weather and Electricity, our model achieves remarkable improvements. For instance, on the ETTm2 dataset (horizon 96), DPWMixer reduces MSE by 10.7\% compared to iTransformer (0.169 vs. 0.180) and by 6.50 \% compared to TimeMixer (0.169 vs. 0.175). These results validate our hypothesis that: the lossless Haar wavelet decomposition effectively preserves high-frequency details (e.g., traffic peaks) that are often smoothed out by the average pooling used in TimeMixer, while the dual-path mixer captures local non-linear dynamics better than the rigid attention mechanisms in Transformers.

Furthermore, on datasets with strong global trends but significant noise, such as Exchange and Weather, DPWMixer continues to lead. In the Exchange dataset, our model achieves an average MSE of 0.171 (horizon 192), significantly outperforming iTransformer (0.177) and TimeMier (0.187). These gains stem from the Global Linear Path in our architecture, which acts as a stable anchor for macroscopic trends, preventing the model from overfitting to aleatoric uncertainty—a common pitfall for deep models like iTransformer in financial data.

Notably, DPWMixer still maintains its performance gap when horizon becomes longer. For instance, on Electricity dataset when horizon reaches to the longest $(T=720)$, our method could obtain MSE=0.223 while the state-of-the-art iTransformer achieves 0.225. It shows that our hierarchical multi-scale design greatly alleviates the error accumulation issue in long-term forecasting. Moreover, the orthogonal wavelet decomposition enables the model to learn long-term trend information in coarser scales independently from high-frequency noisy parts so that the model could be more stable when predicting further into the future.

In summary, the experimental results have shown that, due to the orthogonal wavelet decomposition used for anti-aliasing and dual-path mixer used for harmonizing rigid and flexible modeling, DPWMixer sets a new state-of-the-art accuracy and robustness benchmark for long-term time series forecasting.

\subsection{Visualized prediction results}
To provide a more intuitive understanding of the model's behavior beyond aggregated metrics, we visualize the forecasting results of DPWMixer on Electricity dataset and other seven representative baselines. The Electricity dataset shows non-stationary dynamics and highly periodic behavior with both sharp cyclic changes and high-frequency fluctuations, which makes it an ideal dataset to evaluate the temporal fidelity of models. We visualize the forecasting results of DPWMixer on horizon $T=192$ and $T=336$ respectively with look-back window $L=96$.

\begin{figure}[htbp]
    \centering 
    \begin{minipage}[b]{0.48\columnwidth}
        \centering
        \includegraphics[width=\linewidth]{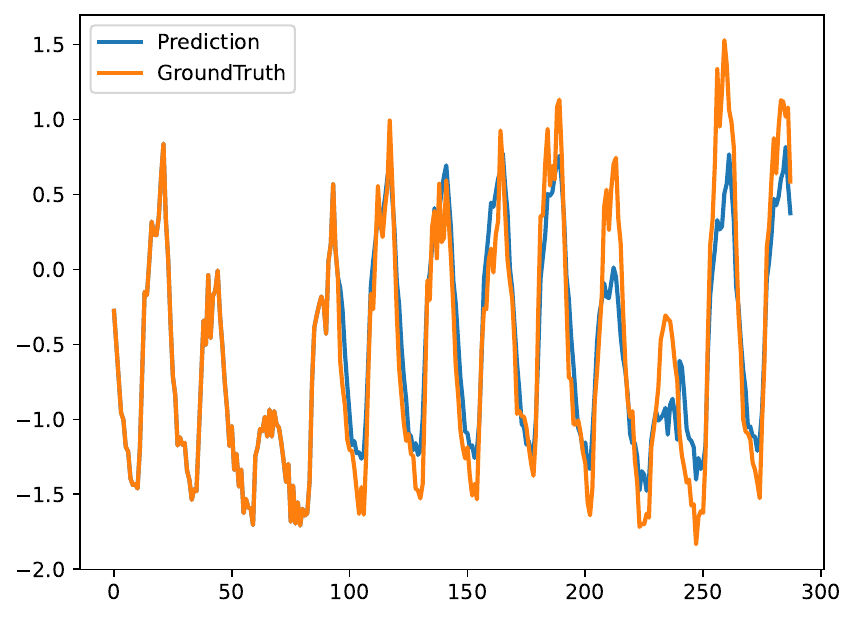}
        \centerline{(a) Electricity-192-DPWMixer}
    \end{minipage}%
    \hfill
    \begin{minipage}[b]{0.48\columnwidth}
        \centering
        \includegraphics[width=\linewidth]{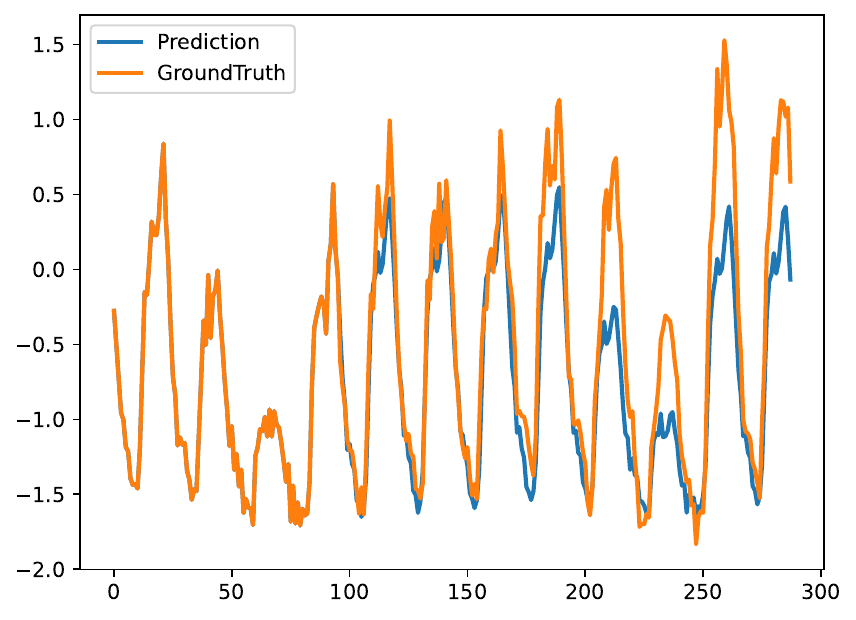}
        \centerline{(b) Electricity-192-iTransformer}
    \end{minipage}
\vspace{0.3cm} 
    \begin{minipage}[b]{0.48\columnwidth}
        \centering
        \includegraphics[width=\linewidth]{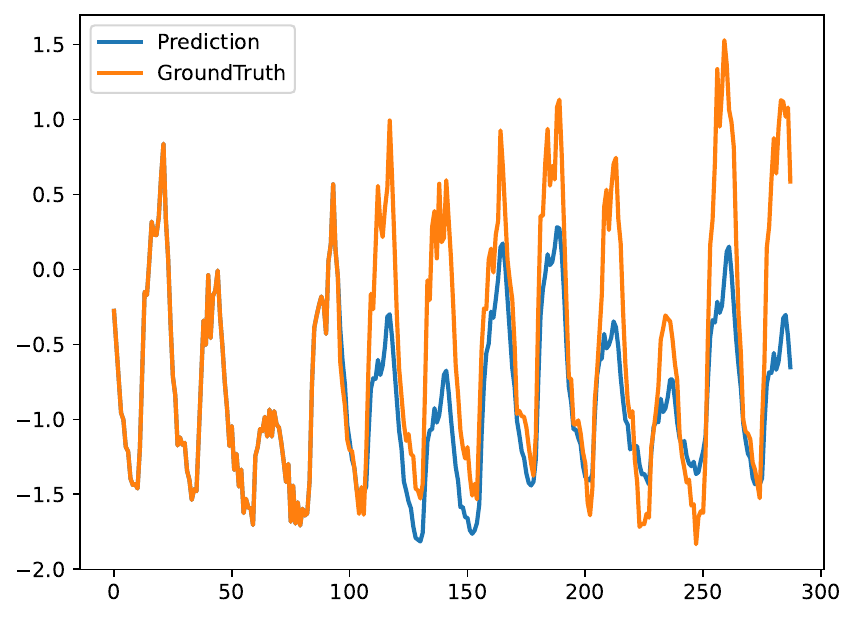}
        \centerline{(c) Electricity-192-TimeMixer}
    \end{minipage}%
    \hfill
    \begin{minipage}[b]{0.48\columnwidth}
        \centering
        \includegraphics[width=\linewidth]{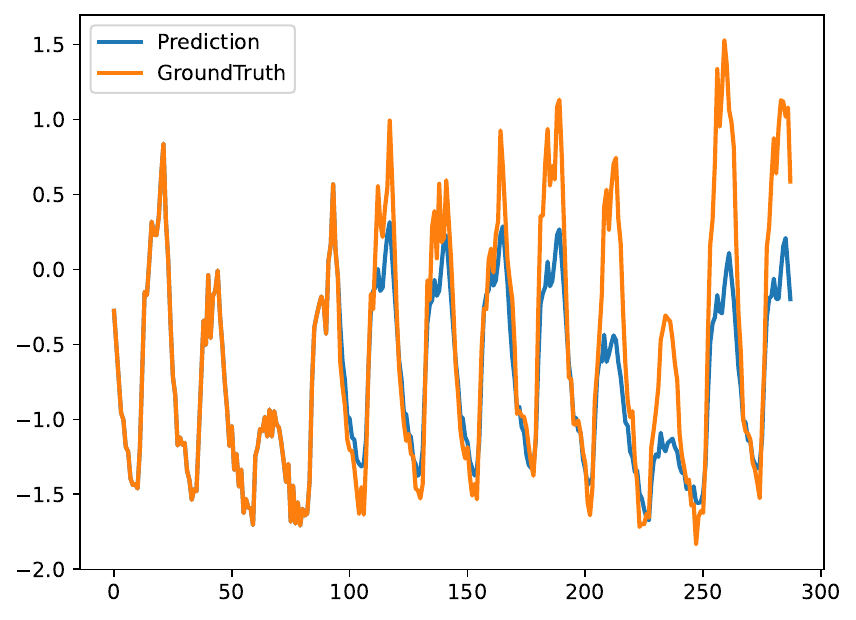}
        \centerline{(d) Electricity-192-TimeXer}
    \end{minipage}
\vspace{0.3cm} 

    \begin{minipage}[b]{0.48\columnwidth}
        \centering
        \includegraphics[width=\linewidth]{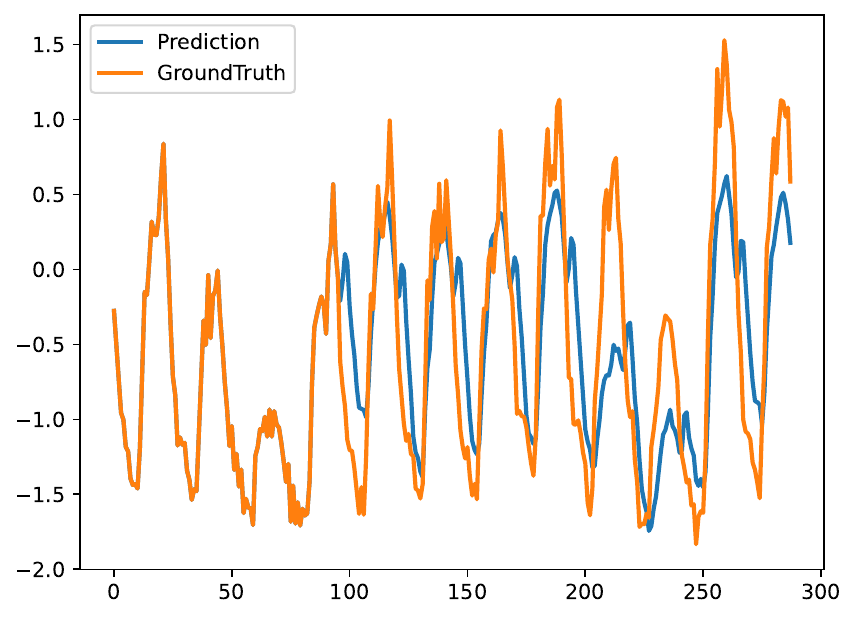}
        \centerline{(e) Electricity-192-CrossFormer}
    \end{minipage}%
    \hfill
    \begin{minipage}[b]{0.48\columnwidth}
        \centering
        \includegraphics[width=\linewidth]{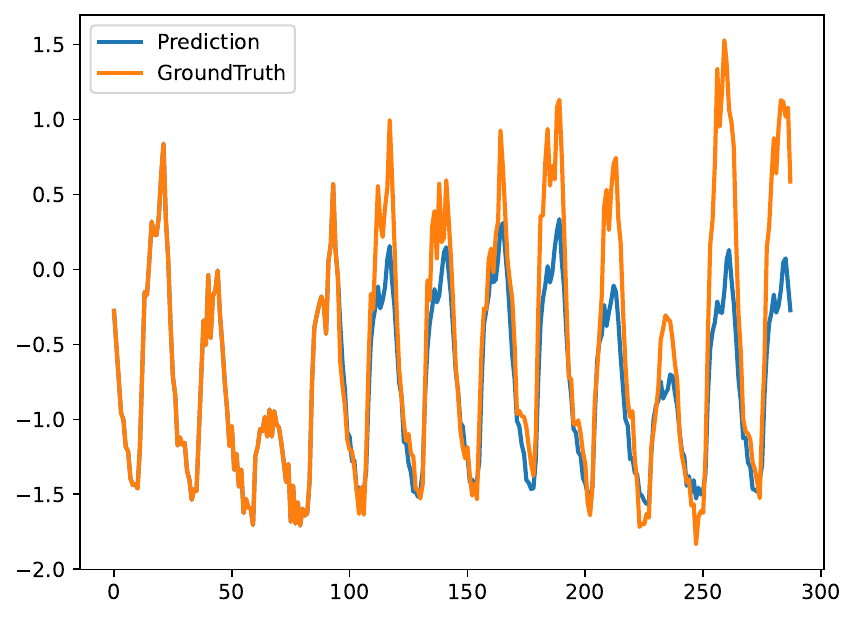}
        \centerline{(f) Electricity-192-PatchTST}
    \end{minipage}
    \vspace{0.3cm} 

    \begin{minipage}[b]{0.48\columnwidth}
        \centering
        \includegraphics[width=\linewidth]{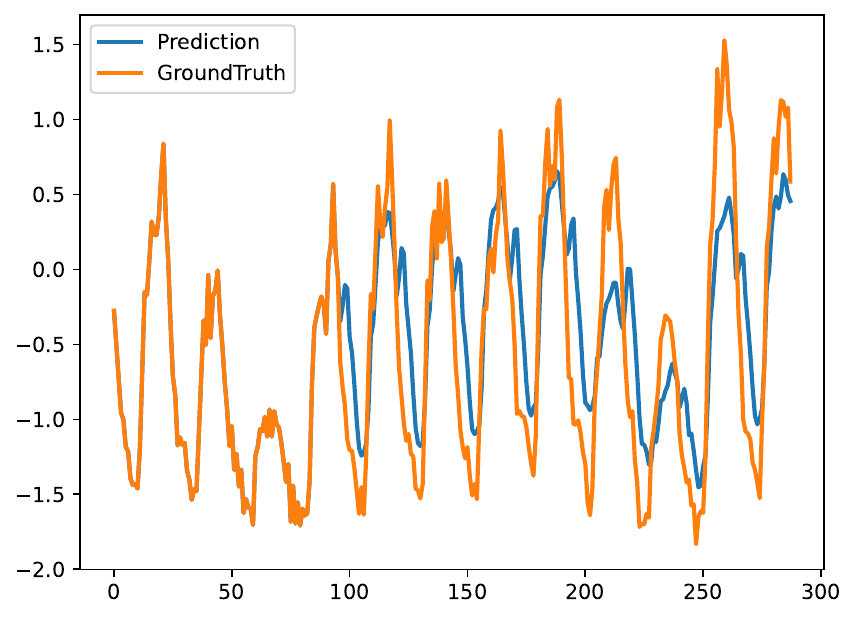}
        \centerline{(g) Electricity-192-TimesNet}
    \end{minipage}%
    \hfill
    \begin{minipage}[b]{0.48\columnwidth}
        \centering
        \includegraphics[width=\linewidth]{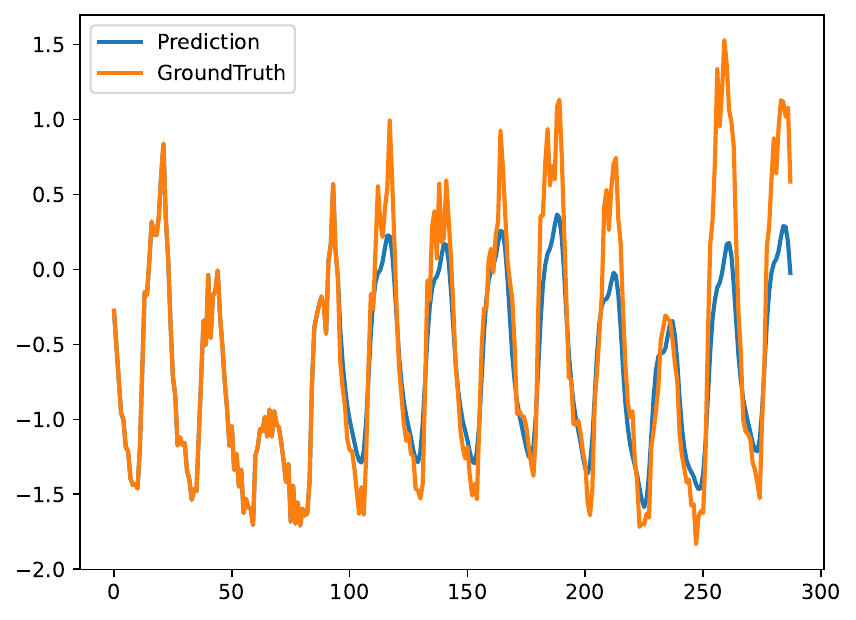}
        \centerline{(h) Electricity-192-Dlinear}
    \end{minipage}
  \caption{Visual comparison of forecasting performance on the Electricity dataset with a horizon of $T=192$. DPWMixer (a) accurately captures cyclic patterns and maintains trend consistency over the extended horizon, exhibiting lower error accumulation than the baseline method.}
    \label{fig:elec_192}
\end{figure}

\textbf{Performance on Horizon $T=192$ (Figure \ref{fig:elec_192}):}
In the medium-term prediction scenario, DPWMixer is almost aligned with the ground truth (Orange line). As shown in Figure \ref{fig:elec_192} (a), our model could reconstruct the fine-grained temporal structure information, which is the sharp peaks and deep troughs of peak electricity load. Compared with other advanced models which could also capture the general periodicity of the time series, DPWMixer could further model the amplitude of extremum points more accurately. Due to the self-attention mechanism, the Transformer-based model would bring a smoothing effect on the original time series, which would lead to under-estimation in high volatility regions. While the Dual-Path design makes Local Evolution Path could focus on reconstructing these high-frequency information independently from global trend information, and thus it would not be smoothed down by other trend components.

\begin{figure}[htbp]
    \centering 
    \begin{minipage}[b]{0.48\columnwidth}
        \centering
        \includegraphics[width=\linewidth]{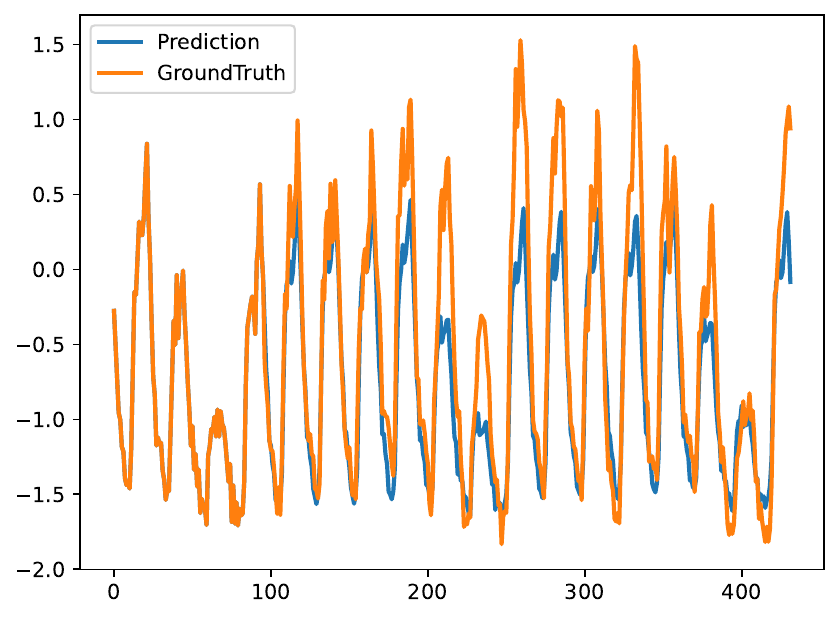}
        \centerline{(a) Electricity-336-DPWMixer}
    \end{minipage}%
    \hfill
    \begin{minipage}[b]{0.48\columnwidth}
        \centering
        \includegraphics[width=\linewidth]{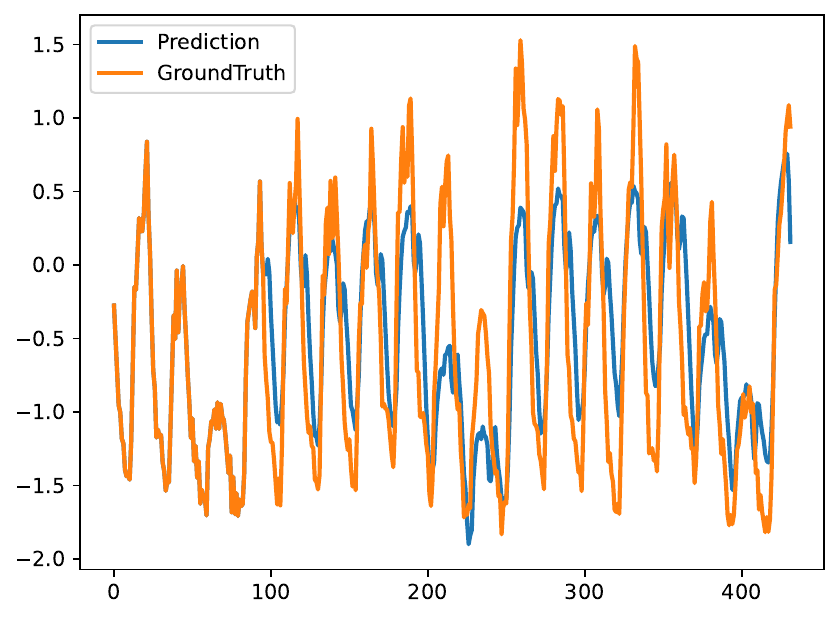}
        \centerline{(b) Electricity-336-iTransformer}
    \end{minipage}
\vspace{0.3cm} 
    \begin{minipage}[b]{0.48\columnwidth}
        \centering
        \includegraphics[width=\linewidth]{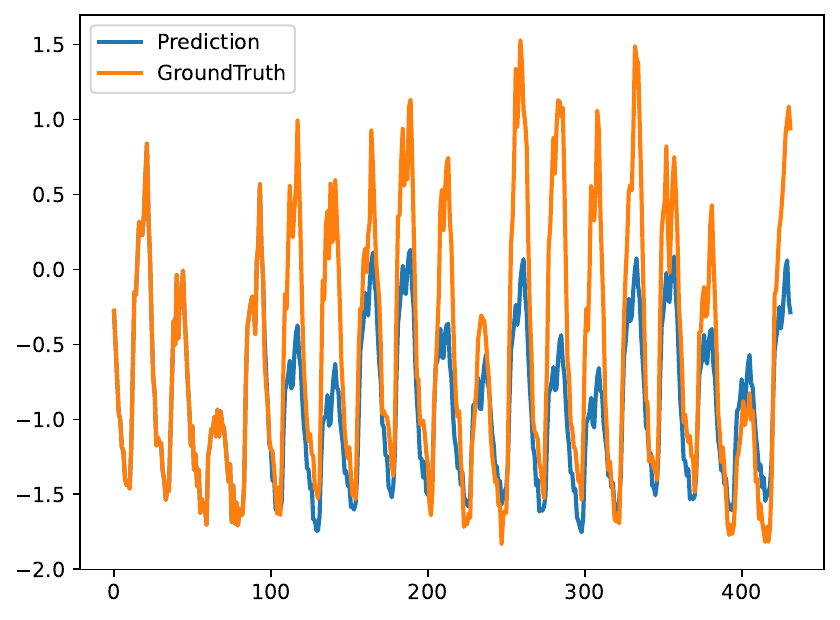}
        \centerline{(c) Electricity-336-TimeMixer}
    \end{minipage}%
    \hfill
    \begin{minipage}[b]{0.48\columnwidth}
        \centering
        \includegraphics[width=\linewidth]{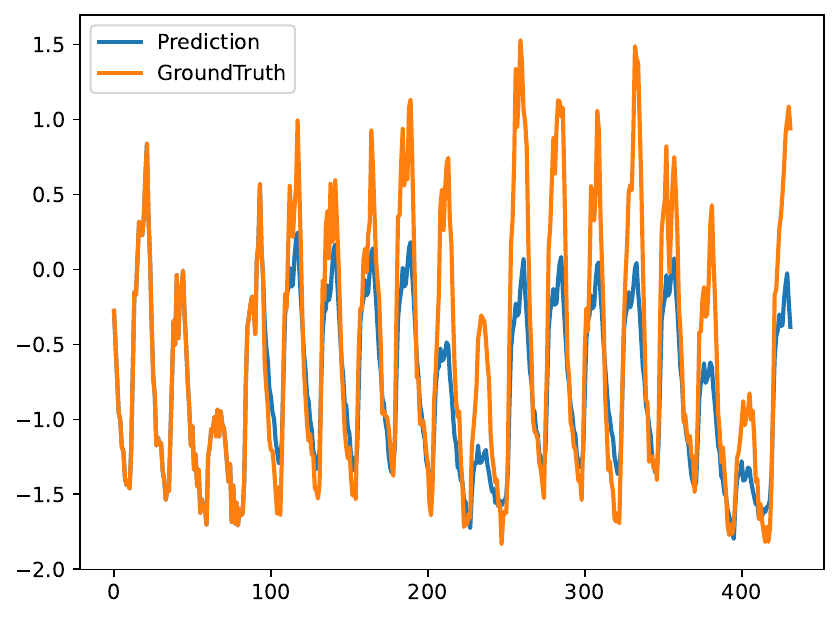}
        \centerline{(d) Electricity-336-TimeXer}
    \end{minipage}
\vspace{0.3cm} 

    \begin{minipage}[b]{0.48\columnwidth}
        \centering
        \includegraphics[width=\linewidth]{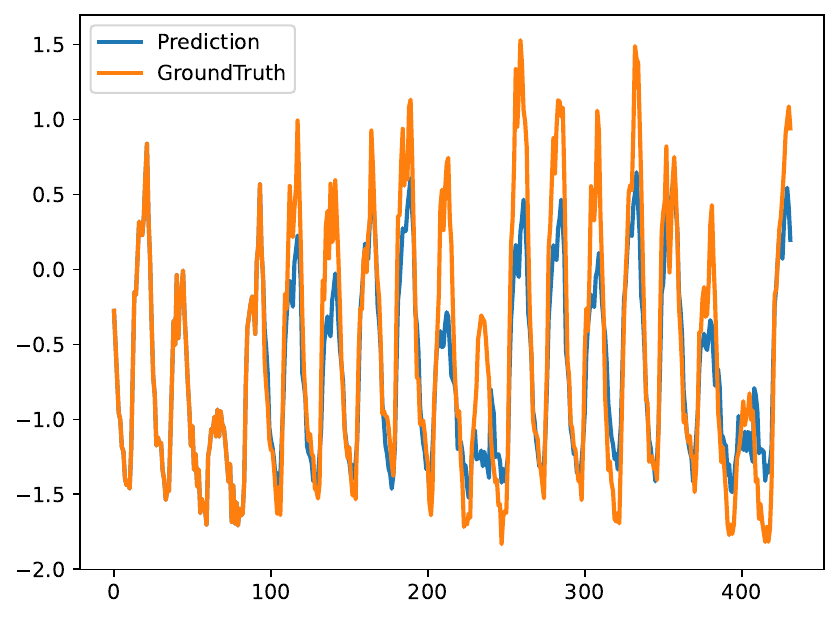}
        \centerline{(e) Electricity-336-CrossFormer}
    \end{minipage}%
    \hfill
    \begin{minipage}[b]{0.48\columnwidth}
        \centering
        \includegraphics[width=\linewidth]{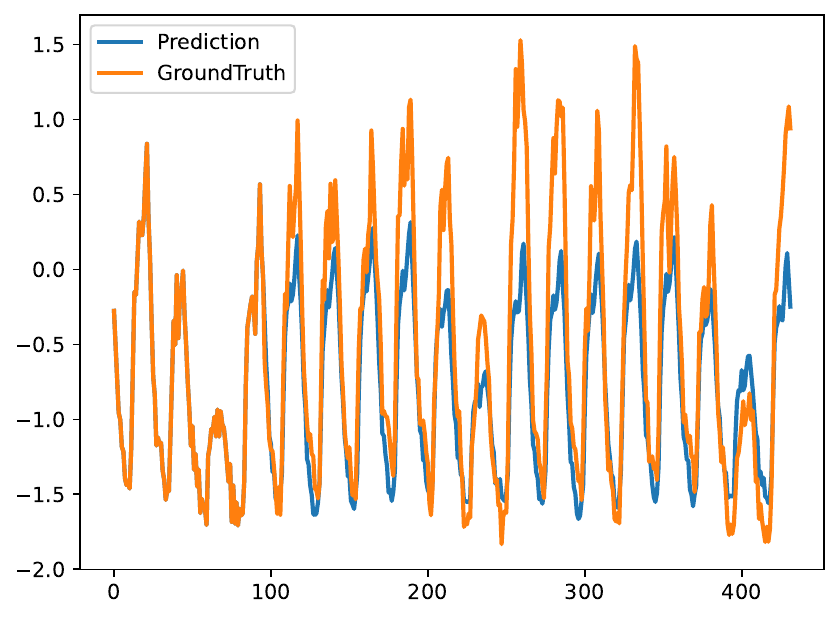}
        \centerline{(f) Electricity-336-PatchTST}
    \end{minipage}
    \vspace{0.3cm} 

    \begin{minipage}[b]{0.48\columnwidth}
        \centering
        \includegraphics[width=\linewidth]{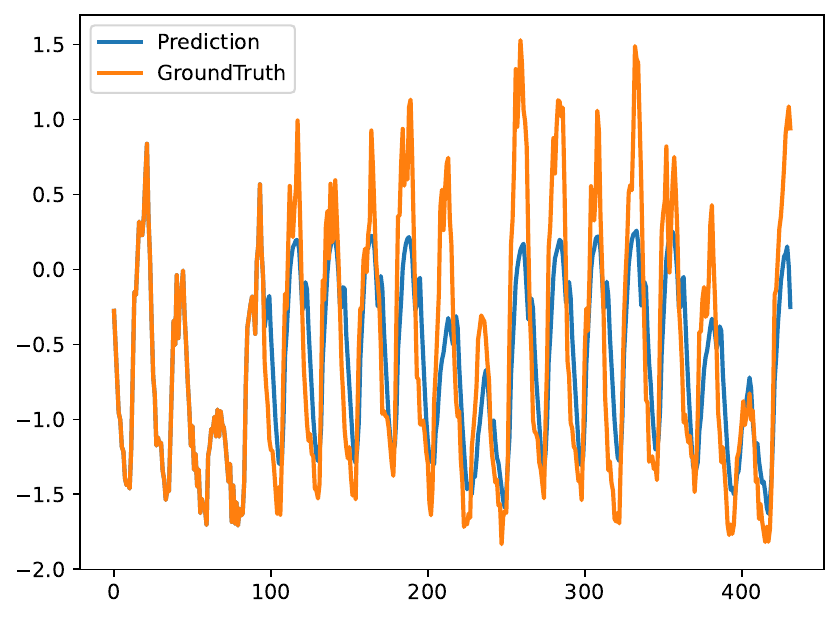}
        \centerline{(g) Electricity-336-TimesNet}
    \end{minipage}%
    \hfill
    \begin{minipage}[b]{0.48\columnwidth}
        \centering
        \includegraphics[width=\linewidth]{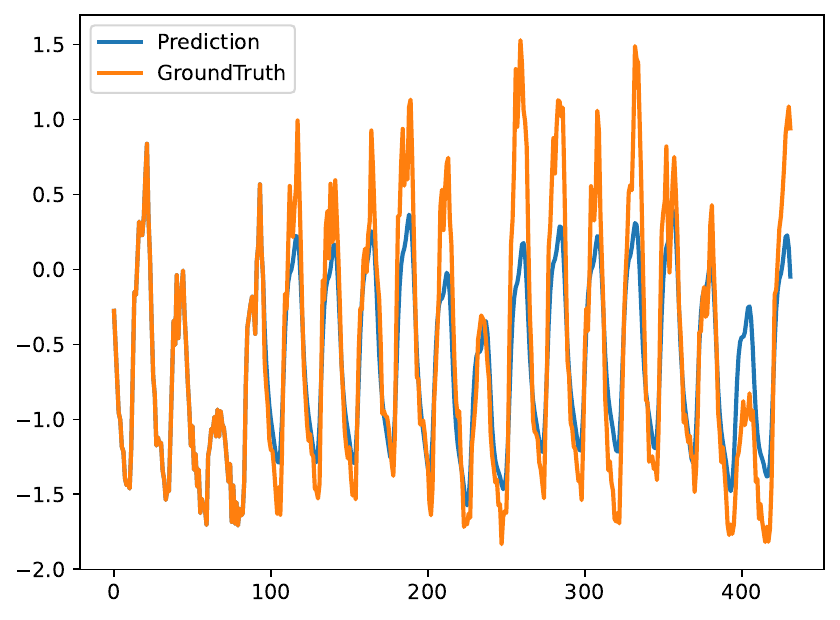}
        \centerline{(h) Electricity-336-Dlinear}
    \end{minipage}
    \caption{Visual comparison of long-term forecasting results on the Electricity dataset ($T=336$). 
DPWMixer (a) accurately captures cyclic patterns and maintains trend consistency over the extended horizon, exhibiting lower error accumulation than the baseline methods.}
     \label{fig:elec_336}
\end{figure}

\textbf{Robustness on Extended Horizon $T=336$ (Figure \ref{fig:elec_336}):}
It is even more apparent that DPWMixer is superior to the other models when $T=336$. It is a challenging task to forecast far into the future because the errors will accumulate over time. It is also a challenge to keep the phase shift stable over a long sequence. It is hard to keep structural consistency when forecasting over a long sequence as shown in Figure \ref{fig:elec_336}. DPWMixer (shown in Figure \ref{fig:elec_336}(a)) is very robust and can keep the tight phase and amplitude consistency over the long sequence and is able to accurately forecast the double-peak patterns.

\textbf{Attribution to Architectural Innovations:}
This visual superiority directly validates the efficacy of our proposed Lossless Haar Wavelet Pyramid. Unlike traditional average pooling, which may incur spectral aliasing and loss of high-frequency information, our orthogonal wavelet decomposition ensures that critical transient details are preserved and passed to the mixer layers. Furthermore, the Global Linear Path provides a stable anchor for the macroscopic trajectory, preventing the trend drift often observed in purely non-linear deep models over long horizons. Consequently, DPWMixer achieves a harmonized balance between trend stability and detail sensitivity, resulting in forecasts that are not only statistically accurate but also structurally faithful to the real-world signal dynamics.

\subsection{Ablation Studies}

To rigorously verify the contribution of each component in DPWMixer, we conduct ablation studies on the Electricity and ETTm1 datasets. We analyze the impact of each component based on the results presented in Table \ref{tab:ablation-study}.

\begin{sidewaystable*}[p]
\caption{Ablation study of DPWMixer components on Electricity and ETTm1 datasets. Ours represents the full model. The degradation row indicates the average percentage increase in MSE compared to the full model.}

\label{tab:ablation-study}
\centering
\resizebox{1.0\textwidth}{!}{
\setlength{\tabcolsep}{0.9em} 
\begin{tabular}{llcccccccccc}
\toprule
\multicolumn{2}{c}{\textbf{Models}} & \multicolumn{2}{c}{\textbf{Ours (Full)}} & \multicolumn{2}{c}{\textbf{w/o Wavelet}} & \multicolumn{2}{c}{\textbf{w/o Global Path}} & \multicolumn{2}{c}{\textbf{w/o Local Path}} & \multicolumn{2}{c}{\textbf{w/o Fusion}} \\
\cmidrule(lr){3-4} \cmidrule(lr){5-6} \cmidrule(lr){7-8} \cmidrule(lr){9-10} \cmidrule(lr){11-12}
\textbf{Dataset} & \textbf{Horizon} & MSE & MAE & MSE & MAE & MSE & MAE & MSE & MAE & MSE & MAE \\
\midrule
\multirow{4}{*}{Electricity} 
& 96  & \textbf{0.152} & \textbf{0.245} & 0.169 & 0.262 & 0.159 & 0.252 & 0.174 & 0.269 & 0.156 & 0.248 \\
& 192 & \textbf{0.160} & \textbf{0.251} & 0.179 & 0.270 & 0.172 & 0.261 & 0.185 & 0.275 & 0.165 & 0.255 \\
& 336 & \textbf{0.173} & \textbf{0.266} & 0.194 & 0.288 & 0.184 & 0.278 & 0.199 & 0.291 & 0.178 & 0.272 \\
& 720 & \textbf{0.223} & \textbf{0.306} & 0.248 & 0.331 & 0.246 & 0.324 & 0.252 & 0.339 & 0.232 & 0.314 \\
\cmidrule(lr){2-12}
\multicolumn{2}{c}{Avg. Degradation} & - & - & +11.8\% & +9.2\% & +7.5\% & +6.8\% & +13.5\% & +10.1\% & +3.2\% & +2.8\% \\
\midrule
\multirow{4}{*}{ETTm1} 
& 96  & \textbf{0.316} & \textbf{0.346} & 0.335 & 0.362 & 0.329 & 0.355 & 0.342 & 0.375 & 0.321 & 0.351 \\
& 192 & \textbf{0.356} & \textbf{0.373} & 0.378 & 0.395 & 0.372 & 0.388 & 0.385 & 0.402 & 0.363 & 0.380 \\
& 336 & \textbf{0.378} & \textbf{0.401} & 0.405 & 0.422 & 0.415 & 0.435 & 0.410 & 0.431 & 0.386 & 0.409 \\
& 720 & \textbf{0.452} & \textbf{0.437} & 0.481 & 0.465 & 0.502 & 0.488 & 0.485 & 0.472 & 0.461 & 0.445 \\
\cmidrule(lr){2-12}
\multicolumn{2}{c}{Avg. Degradation} & - & - & +6.5\% & +5.8\% & +8.9\% & +7.2\% & +8.1\% & +9.5\% & +2.1\% & +2.0\% \\
\bottomrule
\end{tabular}
}
\end{sidewaystable*}

\begin{enumerate}
    \item Impact of Wavelet Decomposition (w/o Wavelet): Replacing the Haar wavelet with average pooling leads to a significant performance drop, particularly on the Electricity dataset (MSE +11.8\%). This confirms that average pooling causes spectral aliasing, losing high-frequency load fluctuation details that are critical for accurate electricity forecasting.
    \item Impact of Global Linear Path (w/o Global Path): Removing the linear path severely impacts long-term horizons. For ETTm1 at $T=720$, the MSE degrades from 0.452 to 0.502. This validates the Global Path acts as a necessary anchor to prevent trend drifting in long sequence predictions.
    \item Impact of Local Mixer Path (w/o Local Path): The removal of the mixer path results in the highest degradation in MAE for Electricity, indicating that the remaining linear component is too rigid to capture the complex, non-linear micro-dynamics of power consumption.
    \item Impact of Adaptive Fusion: While the impact is smaller compared to other modules, the consistent degradation ($\approx 2-3\%$) across all settings shows that statically aggregating multi-scale features is suboptimal compared to our channel-aware dynamic weighting strategy.
\end{enumerate}

\subsection{Parameter Sensitivity Analysis }

\begin{figure}[h!]
    \centering

    \begin{minipage}[b]{0.50\columnwidth}
        \centering
        \includegraphics[width=\linewidth]{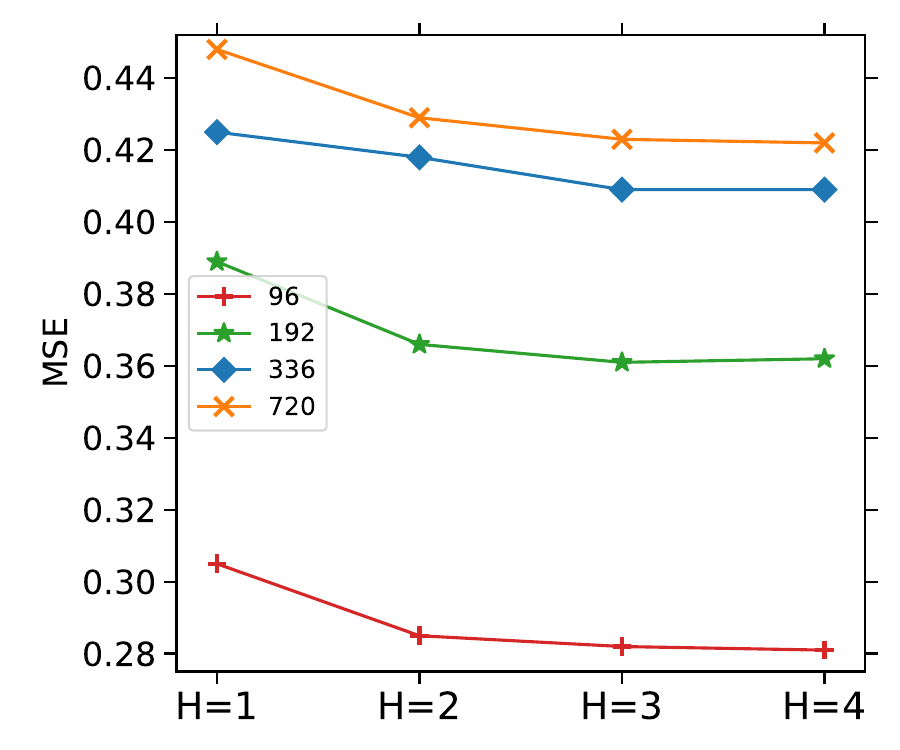}
        \centerline{(a) ETTh2 Dataset}
    \end{minipage}
    \hfill   
    \begin{minipage}[b]{0.50\columnwidth}
        \centering
        \includegraphics[width=\linewidth]{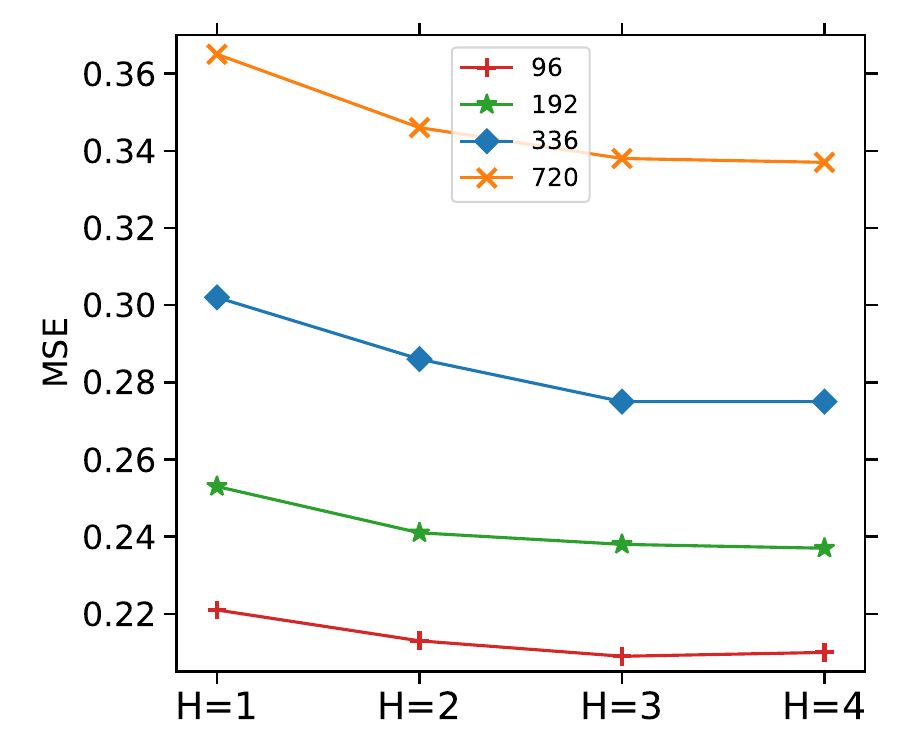}
        \centerline{(b) Weather Dataset}
    \end{minipage}

    \caption{Sensitivity analysis results on ETTh2 and Electricity datasets.}
    \label{fig:sensitivity_n_layers}
\end{figure}

We investigate the impact of the wavelet decomposition depth, denoted as $H$, by varying it from 1 to 4 on the ETTh2 and Weather datasets. As illustrated in Figure \ref{fig:sensitivity_n_layers}, a consistent trend emerges where increasing the decomposition level initially leads to a significant reduction in Mean Squared Error (MSE) across all prediction horizons. The single-scale model ($H=1$) yields the poorest performance, attributed to its limited receptive field and inability to physically disentangle global trends from high-frequency noise. In contrast, deeper hierarchies enable the DPWMixer to capture multi-scale temporal dependencies more effectively, proving that the multi-resolution architecture provides a superior inductive bias for long-term forecasting.

A detailed analysis of individual datasets clarifies the source of these improvements. 
For ETTh2, the significant performance gain observed when increasing $H$ from $1$ to $3$ suggests that coarse-grained coefficients at deeper levels effectively capture long-term trends, mitigating the drift often associated with single-stream models. Similarly, the Weather dataset benefits from multi-scale modeling, where separating high-frequency fluctuations from slow-varying components at $H=2$ and $H=3$ reduces prediction error. In both cases, the hierarchical structure utilizes wavelet orthogonality to preserve signal energy while expanding the temporal receptive field.

However, performance saturates or diminishes at $H=4$. This suggests that excessive down-sampling causes information loss, as the sequence length at the coarsest level (e.g., $L/16$) becomes insufficient to represent complex patterns. Consequently, we select $H=3$ as the optimal setting, as it balances the need for a large receptive field with the retention of local details.

\subsection{Efficiency Analysis}

\begin{figure}[h!]
    \centering
    \includegraphics[width=0.95\columnwidth]{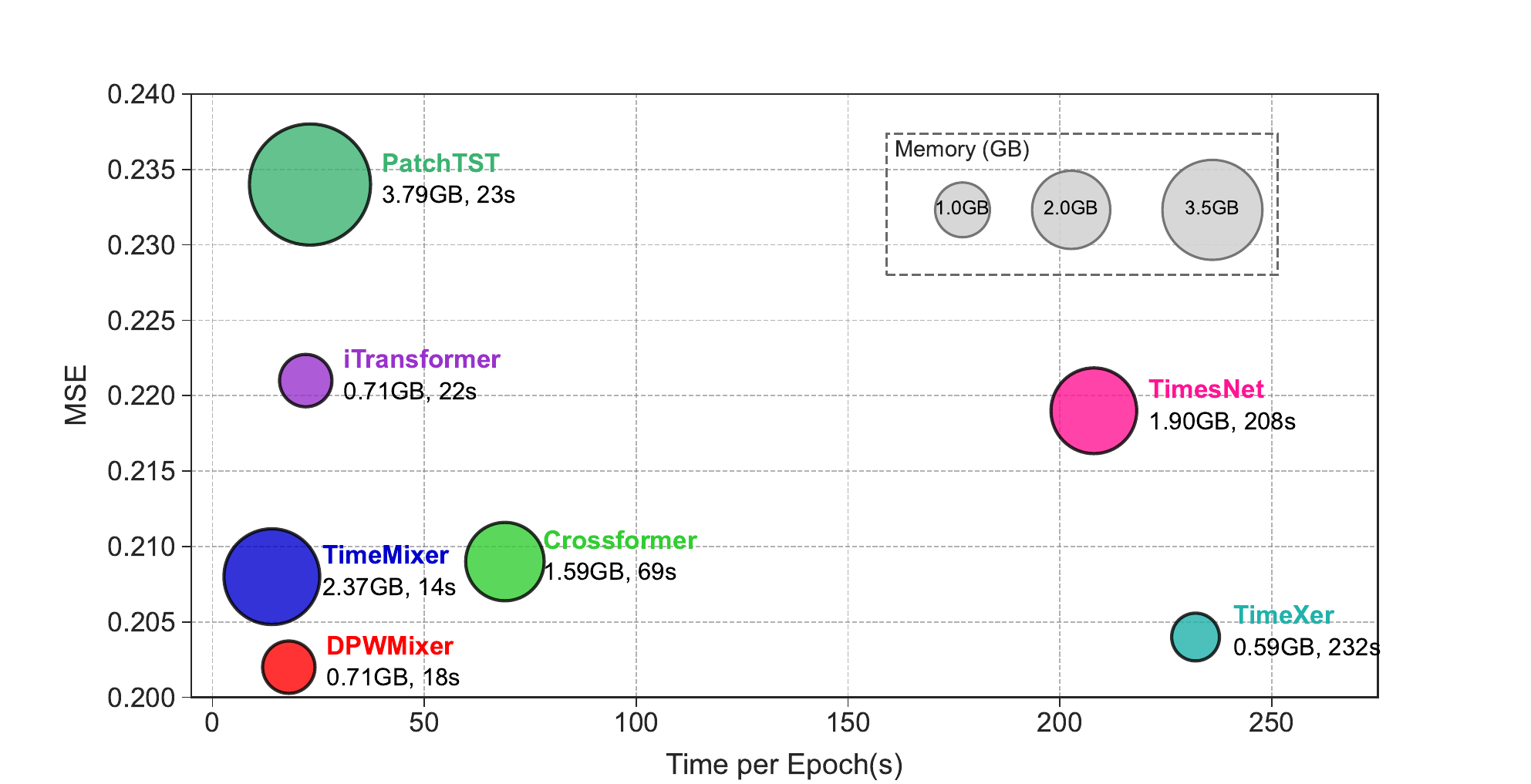} 
    \caption{Efficiency comparison on the weather dataset ($L=96, T=192$). The x-axis denotes training speed (s/epoch), the y-axis denotes MSE, and the bubble area represents GPU memory usage. DPWMixer (Red) achieves the optimal trade-off.}
    \label{fig:efficiency_comparison1}
\end{figure}

\begin{figure}[h!]
    \centering
    \includegraphics[width=0.95\columnwidth]{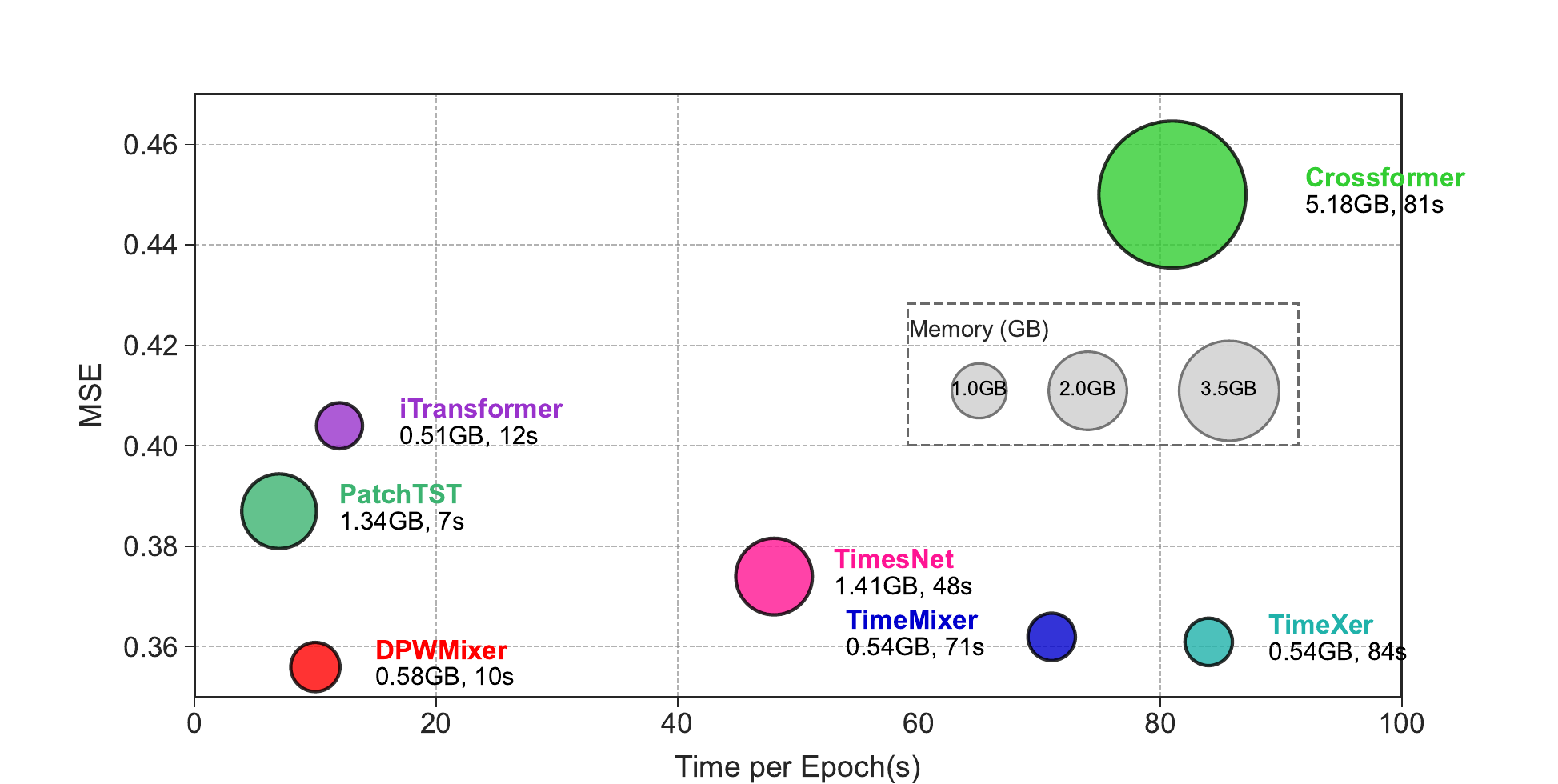} 
    \caption{Efficiency comparison on the ETTm1 dataset ($L=96, T=192$). The x-axis denotes training speed (s/epoch), the y-axis denotes MSE, and the bubble area represents GPU memory usage. DPWMixer (Red) achieves the optimal trade-off.}
    \label{fig:efficiency_comparison2}
\end{figure}

For real-world deployment, the utility of a forecasting model is defined by the trade-off between predictive accuracy and computational cost. We analyze this balance in Figure \ref{fig:efficiency_comparison1} and \ref{fig:efficiency_comparison2}, visualizing a multi-dimensional space where the x-axis represents training speed (time per epoch), the y-axis represents forecasting error (MSE), and the bubble size indicates the GPU memory footprint. Ideally, a model should occupy the bottom-left corner with a minimal bubble size, representing high speed, low error, and memory efficiency. We exclude ultra-lightweight models like DLinear and FITS, as their limited capacity to capture complex non-linear dynamics results in significant accuracy penalties on challenging datasets. Benchmarking high-capacity deep architectures against such simple mappings skews the visualization and obscures the critical trade-off between representation power and computational efficiency.

As illustrated, DPWMixer (represented by the red bubble) consistently achieves the optimal frontier.
\textbf{Superiority over Transformers:} Due to the quadratic complexity $\mathcal{O}(L^2)$ of attention mechanism, the deep architectures like Crossformer and PatchTST are very computationally expensive. While DPWMixer only consumes 0.58GB memory, it achieves 3.25 \% MSE on ETTh1 dataset compared with Crossformer (5.18GB). The efficient Transformer decoding is also applied to our Dual-Path Mixers, which geometrically reduce sequence length to linear $\mathcal{O}(L)$ by Haar Wavelet Pyramid. Our model decomposes the series into multi-scale representations and then models them in the pyramid structure, which means the subsequent Dual-Path Mixers are actually working on much fewer tokens (at the deepest layers, the sequence length is $L/2^N$ where N is the level of the pyramid).

\textbf{Superiority over Efficient Baselines:} Compared with TimesNet, which needs computationally expensive 2D convolutions on spatial and temporal modeling and variance modeling, DPWMixer is 4$\times$ faster. And TimeMixer is designed to be efficient using average pooling, while DPWMixer is even better in accuracy, which shows that our orthogonal wavelet decomposition could preserve more high-frequency details than pooling, and achieves a better trade-off between speed and fidelity.

In conclusion, DPWMixer combines wavelets and dual-path mixing to capture both global trends and local details, achieving top-tier accuracy with high computational efficiency.

\section{Conclusion}\label{con}
In this paper, we propose DPWMixer, a novel and computationally efficient framework for long-term time series forecasting. We identified the spectral aliasing issue in traditional pooling and proposed the Lossless Haar Wavelet Pyramid to achieve orthogonal decomposition of trends and details. Additionally, the proposed Dual-Path Trend Mixer effectively harmonizes rigid global trends and flexible local dynamics. Extensive experiments confirm that DPWMixer significantly outperforms Transformer-based baselines while maintaining linear complexity $\mathcal{O}(L)$, ensuring scalability for high-frequency forecasting tasks. Future work will focus on exploring learnable wavelet transforms to adaptively extract features and scaling this architecture for self-supervised pre-training.

\backmatter

\bmhead*{Acknowledgments}
This work is supported by the National Natural Science Foundation of China (No. 62372366) .

\bmhead*{Data availability}
 Data is available on request.
 
\section*{Declarations}
\textbf{Conflict of interest} The authors declare that they have no known competing financial interests or personal relationships that could have appeared to influence the work reported in this paper.


\bibliography{sn-bibliography}

\end{document}